%% file: main.tex
\documentclass[acmsmall]{acmart}

\AtBeginDocument{%
  \providecommand\BibTeX{{%
    \normalfont B\kern-0.5em{\scshape i\kern-0.25em b}\kern-0.8em\TeX}}}

\usepackage{amsmath,amsfonts}
\usepackage{mathrsfs}
\usepackage{url}

\usepackage{amssymb}
\usepackage{subfigure}
\usepackage[justification=centering]{caption}
\usepackage[ruled,linesnumbered]{algorithm2e}
\usepackage{soul}
\usepackage{xcolor}
\usepackage{bm}
\usepackage{booktabs}
\usepackage{makecell}
\newtheorem{definition}{Definition}
\DeclareMathOperator*{\argmax}{arg\,max}

\begin{document}
\title{Traceable Group-Wise Self-Optimizing Feature Transformation Learning: A Dual Optimization Perspective}

\author{Meng Xiao}
\email{xiaomen7890@gmail.com}
\orcid{0000-0001-5294-5776}
\affiliation{%
  \institution{1.Computer Network Information Center, Chinese Academy of Sciences, Beijing; 2.University of Chinese Academy of Sciences, Beijing}
     % \institution{}
  \country{China}
}

\author{Dongjie Wang}
\affiliation{%
  \institution{Department of Computer Science, University of Central Florida, Orlando}
  \country{USA}}
\email{wangdongjie@knights.ucf.edu}

\author{Min Wu}
\affiliation{%
  \institution{Institute for Infocomm Research, Agency for Science}
  \country{Singapore}}
\email{wumin@i2r.a-star.edu.sg}

\author{Kunpeng Liu}
\email{kunpeng@pdx.edu}
\affiliation{%
   \institution{Department of Computer Science, Portland State University}
  \city{Portland}
  \country{USA}
}

\author{Hui Xiong}
\email{huixiong@ust.hk}
 \affiliation{\institution{1.The Hong Kong University of Science and Technology (Guangzhou); 2. Guangzhou HKUST Fok Ying Tung Research Institute} \country{China}}
 
\author{Yuanchun Zhou}
\email{zyc@cnic.cn}
\affiliation{%
   \institution{Computer Network Information Center, Chinese Academy of Sciences; 2.University of Chinese Academy of Sciences, Beijing}
  \city{Beijing}
  \country{China}
}

\author{Yanjie Fu}
\affiliation{%
  \institution{Department of Computer Science, University of Central Florida, Orlando}
  \country{USA}}

\email{yanjie.fu@ucf.edu}
\thanks{Yanjie Fu and Yuanchun Zhou are the corresponding authors.}
\thanks{Meng Xiao and Dongjie Wang contributed equally to this work.}
\renewcommand{\shortauthors}{Meng and Dongjie, et al.}

\begin{abstract}
Feature transformation aims to reconstruct an effective representation space by mathematically refining the existing features. It serves as a pivotal approach to combat the curse of dimensionality, enhance model generalization, mitigate data sparsity, and extend the applicability of classical models. Existing research predominantly focuses on domain knowledge-based feature engineering or learning latent representations. However, these methods, while insightful, lack full automation and fail to yield a traceable and optimal representation space. An indispensable question arises: Can we concurrently address these limitations when reconstructing a feature space for a machine learning task?
Our initial work took a pioneering step towards this challenge by introducing a novel self-optimizing framework. This framework leverages the power of three cascading reinforced agents to automatically select candidate features and operations for generating improved feature transformation combinations. Despite the impressive strides made, there was room for enhancing its effectiveness and generalization capability.
In this extended journal version, we advance our initial work from two distinct yet interconnected perspectives:
1) We propose a refinement of the original framework, which integrates a graph-based state representation method to capture the feature interactions more effectively and develop different Q-learning strategies to alleviate Q-value overestimation further.
2) We utilize a new optimization technique (actor-critic) to train the entire self-optimizing framework in order to accelerate the model convergence and improve the feature transformation performance. 
Finally, to validate the improved effectiveness and generalization capability of our framework, we perform extensive experiments and conduct comprehensive analyses. These provide empirical evidence of the strides made in this journal version over the initial work, solidifying our framework's standing as a substantial contribution to the field of automated feature transformation.
To improve the reproducibility, we have released the associated code and data by the Github link~\url{https://github.com/coco11563/TKDD2023_code}.

\end{abstract}

\maketitle

\input{introduction}

\input{preliminary}

\input{method}

\input{experiment}

\input{related}

\input{conclusion}

\input{acknowledgement}

\bibliographystyle{ACM-Reference-Format}
\bibliography{Yanjie, acm, meng}

\end{document}

%% file: introduction.tex
% \vspace{-0.1cm}
\section{Introduction}

Classic Machine Learning (ML) mainly includes data prepossessing, feature extraction, feature engineering, predictive modeling, and evaluation. 
%Much efforts have been devoted to developing sophisticated learning models, including, approximation function, error measurement, training algorithms, performance bounds. 
%The parameter numbers of deep models have increased from million level to trillion level in the past years~\cite{fedus2021switch}. 
The evolution of deep AI, however, has resulted in a new principled and widely used paradigm: i) collecting data, ii) computing data representations, and iii) applying ML models. 
Indeed, the success of ML algorithms highly depends on
data representation~\cite{bengio2013representation}. 
%This is because different representations (or features) can entangle and hide more or less the different explanatory factors of variation behind the data.
Building a good representation space is critical and fundamental because it can help to 1) identify and disentangle the underlying explanatory factors hidden in observed data, 2)  easy the extraction of useful information in predictive modeling,  3) reconstruct distance measures to form discriminative and machine-learnable patterns, 4) embed structure knowledge and priors into representation space and thus make classic ML algorithms available to complex graph, spatiotemporal, sequence, multi-media, or even hybrid data. 

\begin{figure}[!htbp]
    \centering
    \vspace{-0.3cm}
    \includegraphics[width=0.65\linewidth]{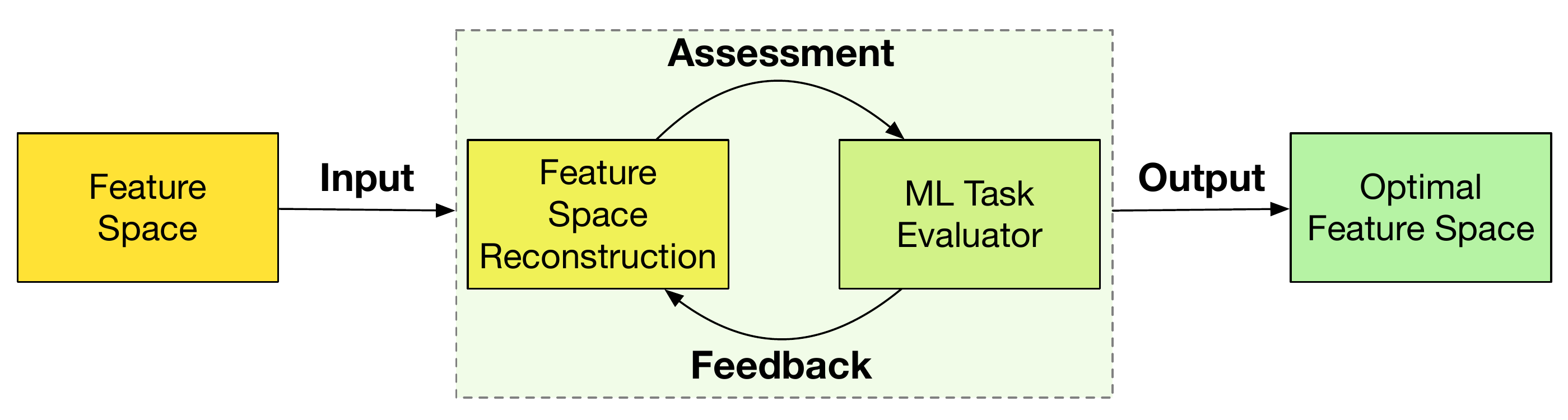}
    \captionsetup{justification=centering}
    \vspace{-0.3cm}
    \caption{We want to uncover the optimal feature space that is traceable and performs optimally in a downstream ML task by iteratively reconstructing the feature space.
    }
    \vspace{-0.3cm}
    \label{fig:intro_frame}
\end{figure}

%representation learning, i.e., learning transformations of the data that make it easier to extract useful information when building classifiers or other predictors. 
In this paper, we study the problem of  learning to reconstruct an optimal and traceable feature representation space to advance a downstream ML task (\textbf{Figure \ref{fig:intro_frame}}).
Formally, given a set of original features, a prediction target, and a downstream ML task (e.g., classification, regression), the objective is to automatically reconstruct an optimal and traceable set of features for the ML task by mathematically transforming original features.

Prior literature has partially addressed the problem. 
The first relevant work is feature engineering, which designs preprocessing, feature extraction, selection~\cite{li2017feature,guyon2003introduction}, and generation~\cite{khurana2018feature} to extract a transformed representation of the data. These techniques are essential but labor-intensive, showing the low applicability of current ML practice in the automation of extracting a discriminative feature representation space.
\textbf{Issue 1 (full automation): } \emph{how can we make ML less dependent on feature engineering, construct ML systems faster, and expand the scope and applicability of ML?}
The second relevant work is representation learning, such as factorization~\cite{fusi2018probabilistic},  embedding~\cite{goyal2018graph}, and deep representation learning~\cite{wang2021reinforced,wang2021automated}. 
These studies are devoted to learning effective latent features. However, the learned features are implicit and non-explainable.
Such traits limit the deployment of these approaches in many application scenarios (e.g., patient and biomedical domains) that require not just high predictive accuracy but also trusted understanding and interpretation of underlying drivers. 
\textbf{Issue 2 (explainable explicitness): }
\emph{how can we assure that the reconstructing representation space is traceable and explainable?}
The third relevant work is learning based feature transformation, such as  principle component analysis~\cite{candes2011robust}, traversal transformation graph based feature generation~\cite{khurana2018feature}, sparsity regularization based feature selection~\cite{hastie2019statistical}.
These methods are either deeply embedded into or totally irrelevant to a specific ML model. 
For example, LASSO regression extracts an optimal feature subset for regression, but not for any given ML model. 
\textbf{Issue 3 (flexible optimal): } \emph{how can we create a framework to reconstruct a new representation space for any given predictor?} 
The three issues are well-known challenges. Our goal is to develop a new perspective to address these issues. 

In the preliminary work~\cite{wang2022multi}, we propose a novel and  principled  \textbf{Traceable Group-wise Reinforcement Generation Framework} for addressing the automation, explicitness, and  optimal issues in representation space reconstruction. 
Specifically, we view feature space reconstruction as a traceable iterative process of the combination of feature generation and feature selection. The generation step is to make new features by mathematically transforming the original features, and the selection step is to avoid the feature set explosion.
To make the process self-optimizing, we formulate it into three Markov Decision Processes (MDPs): one is to select the first meta feature, one is to select an operation, and one is to select the second meta feature.
The two candidate features are transformed by conducting the selected operation on them to generate new ones.
We develop a cascading agent structure to coordinate the three MDPs and learn better feature transformation policies.
Meanwhile, we propose a group-wise feature generation mechanism to accelerate feature space reconstruction and augment reward signals in order for cascading agents to learn clear policies.

While our initial research has achieved significant results, there remains scope for further improvement. 
To improve the performance and generalization capabilities of our initial work, we explore two distinct yet interconnected avenues:
1) Refinement of the existing framework: We propose a graph-based state representation method, enabling reinforced agents to perceive the current feature space more effectively. Furthermore, we employ enhanced Q-learning strategies to mitigate the overestimation of Q-values during policy learning.
2) Introduction of a novel optimization approach: We utilize an Actor-Critic policy learning method to train the overall framework. It aims to expedite model convergence and enhance feature transformation performance.

% {\color{red} Even though the preliminary work has done well, the effectiveness of the reconstructed feature space can still be enhanced by: \textbf{(1): obtaining enhanced state representation}, which enables reinforced agents to comprehend the current feature set better; \textbf{(2): addressing Q-value overestimation}, which leads reinforced agents to learn unbiased and effective policies. Thus, in this journal version, we provide strategies to meet the two objectives.}

In the first avenue, we assume that a robust state representation can empower reinforced agents to accurately comprehend the current feature space, thereby facilitating the learning of more intelligent policies for feature transformation. 
Consequently, for the graph-based state representation method, we initially construct a comprehensive attributed graph predicated on feature-feature correlations. 
Each node signifies a feature, each edge weight denotes the correlation between two nodes, and each node attribute corresponds to the feature vector. 
Then, we capture the knowledge of the entire feature set and the interactions between features via message passing and graph reconstruction, preserving this information in an embedding vector as a state representation.
Besides, we recognize Q-value overestimation as a critical challenge that hinders the unbiased learning of transformation policies, subsequently leading to sub-optimal model performance. Thus, to mitigate this issue, we employ Q-learning using a dual network structure: one network for action selection, and a second for Q-value estimation. Importantly, these two networks are not updated concurrently. This asynchronous update strategy successfully decouples the max operation in traditional Q-learning, thereby reducing Q-value overestimation and enabling the learning of more robust  policies.

In the second avenue,  we examined the optimization technique of the original framework and observed that Q-learning treats all candidate features as the complete action space. It results in substantial computational costs when our framework deals with high-dimensional spaces. Simultaneously, the $\epsilon$-greedy-based learning strategy of Q-learning presents challenges in enabling reinforced agents to balance the trade-off between efficiency and effectiveness. Consequently, we utilize the Actor-Critic technique to optimize the entire feature transformation framework in order to directly learn intelligent feature-feature crossing policies.
In particular, we develop three Actor-Critic agents to select candidate features and operations for feature transformation. 
 The actor component directly learns the selection policy through policy gradient update and makes the action selection by sampling.
The critic component leverages gradient signals from  temporal-difference errors at each iteration to evaluate and improve selection policies. 
The collaborated relationships between the actor and critic components can expedite model convergence and generate effective  combinations for feature transformation.

In summary, in this paper, we develop a generic and principled framework: group-wise reinforcement feature transformation, for optimal and explainable representation space reconstruction.
The contributions of this paper are summarized as follows:
\begin{enumerate}
    \item We formulate the feature space reconstruction as an interactive process of nested feature generation and selection, where feature generation is to generate new meaningful and explicit features, and feature selection is to remove redundant features to control feature sizes. 
    \item We develop a group-wise reinforcement feature transformation framework to automatically and efficiently generate an optimal and traceable feature set for a downstream ML task without much professional experience and human intervention.
    \item We present refinements for the original framework. In particular, a graph-based state representation method is proposed to capture the feature-feature correlations and the knowledge of the entire feature set for improving the comprehensive of reinforced agents. 
    A decoupling Q-learning strategy is employed to alleviate the Q-value overestimation for learning unbiased and effective policies.
    \item We utilize the Actor-Critic optimization technique to reimplement the self-optimizing feature transformation framework.
    The actor component directly learns the selection policies for candidate features and operations instead of finding a surrogate value function. 
    The critic component evaluates the selection policy and provides feedback to the actor in order to make it learn better policies at the next iteration.
    \item Extensive experiments and case studies are conducted to demonstrate the efficacy of our method. Compared with the preliminary work, we add the outlier detection task with five additional datasets, evaluate different state representation approaches, compare different Q-learning strategies and optimization techniques, and provide comprehensive  analyses.

\end{enumerate}

%% file: preliminary.tex
% \vspace{-0.1cm}
\section{Definitions and Problem Statement}
% In this section, we present several important definitions and then outline the problem statement.

\subsection{Important Definitions}
\begin{definition}
\textbf{Feature Group.} 
We aim to reconstruct the feature space of such datasets $\mathcal{D}<\mathcal{F},y>$. 
Here, $\mathcal{F}$ is a feature set, in which each column denotes a feature and each row denotes a data sample;
$y$ is the target label set corresponding to samples.
To effectively and efficiently produce new features, we divide the feature set $\mathcal{F}$ into different feature groups via clustering, denoted by $\mathcal{C}$.
Each feature group  is a feature subset of  $\mathcal{F}$.
% The intra-cluster features are similar in terms of information distribution, however the inter-cluster features are distinct.

\end{definition}

% \begin{definition}
% \textbf{Operation Set.}
% To effectively and efficiently produce new features, we divide the feature set $\mathcal{F}$ into different feature clusters denoted by $\mathcal{C}$.
% Each $\mathcal{C}$ is the feature subset of $\mathcal{F}$.
% The intra-cluster features are similar in terms of information distribution, however the inter-cluster features are distinct.
% \end{definition}

\begin{definition}
\textbf{Operation Set.}
We perform a mathematical operation on existing features in order to generate new ones.
The collection of all operations is an operation set, denoted by $\mathcal{O}$.
There are two types of operations: unary and binary.
The unary operations include ``square'', ``exp'', ``log'', and etc.
The binary operations are ``plus'', ``multiply'', ``divide'', and etc.
\end{definition}

\begin{definition}
\textbf{Cascading Agent.}
To address the feature generation challenge, we develop a new cascading agent structure. 
This structure is made up of three agents: two feature group agents and one operation agent. 
Such cascading agents share state information and sequentially select feature groups and operations.
\end{definition}
\vspace{-0.5cm}
% \begin{definition}
% \textbf{Feature Interaction.}
% We generate new features through group-group feature interaction.
% To be precise, agents can select one operator and two feature clusters.
% If the operator is unary, we will apply it to all features of the feature cluster that is more relevant to the target label.
% If the operator is binary, we will apply it to the topK dissimilar feature pairs between the two clusters.

% \end{definition}

% \begin{definition}
% \textbf{Downstream Task.}
% We assess the feature set's quality by performing a downstream task such as regression or classification.
% The quantitative  indicator (\textit{e.g.} 1-relative absolute error or f1) of the task is the reflection of the quality.
% The higher of the value of such an indicator, the better the feature set's quality.
% \end{definition}

\begin{figure*}[!t]
    \centering
    \includegraphics[width=1.0\linewidth]{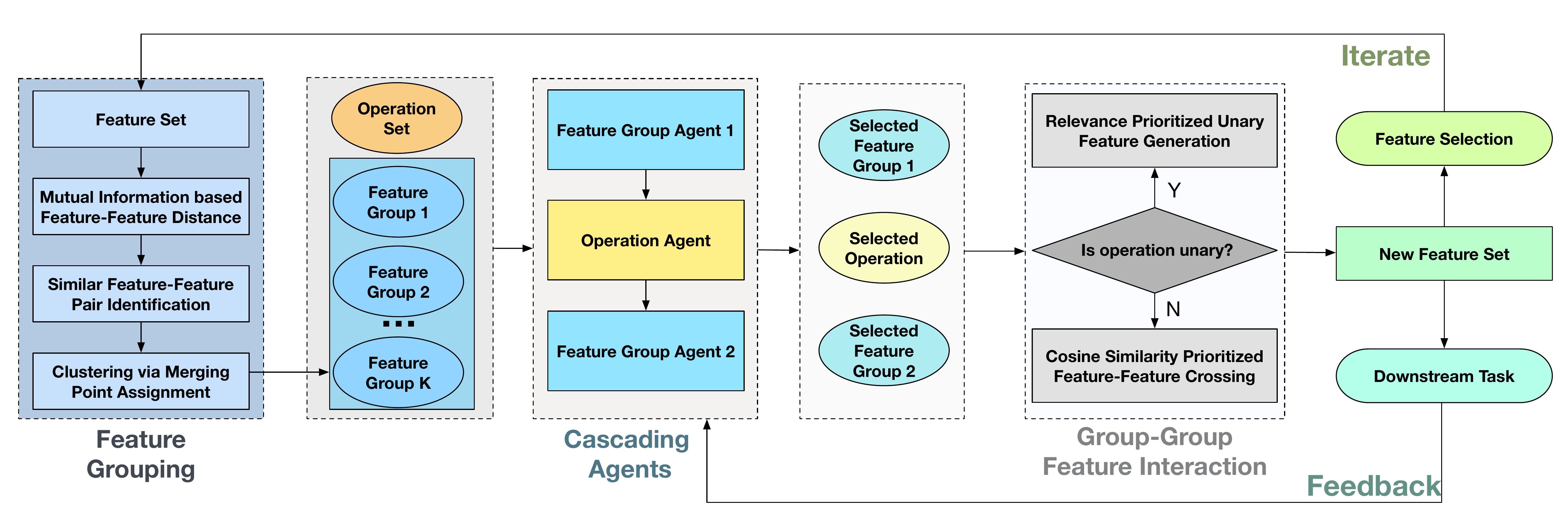}
    \vspace{-0.2cm}
    \captionsetup{justification=centering}
    \vspace{-0.33cm}
    \caption{An overview of the proposed framework GRFG. Feature grouping is to cluster similar features for group-wise feature generation.
    Cascading agents aim to select candidate feature groups and mathematical operations.
    Group-Group feature interaction is to generate new features by crossing selected feature groups.
    }
    \vspace{-0.4cm}
    \label{fig:framework}
\end{figure*}

\subsection{Problem Statement}
The research problem is learning to reconstruct an optimal and explainable feature representation space to advance a downstream ML task.
Formally, given a dataset $D<\mathcal{F},y>$ that includes an original feature set $\mathcal{F}$ and a target label set $y$, an operator set $\mathcal{O}$,   and a downstream ML task $A$ (e.g., classification, regression, ranking, detection), our objective is to automatically reconstruct an optimal and explainable feature set $\mathcal{F}^{*}$   that optimizes the performance indicator $V$ of the task $A$. 
The optimization objective  is to find a reconstructed feature set $\mathcal{\hat{F}}$ that maximizes:
\begin{equation}
\label{objective}
    \mathcal{F}^{*} = argmax_{\mathcal{\hat{F}}}( V_A(\mathcal{\hat{F}},y)),
\end{equation}
where $\mathcal{\hat{F}}$ can be viewed as a subset of a combination of  the original feature set $\mathcal{F}$ and the generated new features $\mathcal{F}^g$, and $\mathcal{F}^g$ is produced by applying the operations $\mathcal{O}$ to  the original feature set $\mathcal{F}$ via a certain algorithmic structure.
%so $\mathcal{\hat{F}} \subseteq  \mathcal{F} \bigcup \mathcal{F}^g$, where $\mathcal{F}^g$ is produced by applying $\mathcal{O}$ to $\mathcal{F}$.

\iffalse
The purpose of this research is to present a framework for automated feature generation.
This framework is capable of autonomously generating relevant features by performing mathematical operators on existing feature sets until the optimal feature set is discovered.
Formally, given an operator set $\mathcal{O}$ and a dataset $D<\mathcal{F},y>$ constituted of a feature set $\mathcal{F}$ and a target label set $y$,  we aim to seek the optimal feature set $\mathcal{F}^{*}$ that optimizes the performance indicator $V$ of the downstream task $A$. 
The objective can be formulated as:
\begin{equation}
\label{objective}
    \mathcal{F}^{*} = argmax_{\mathcal{\hat{F}}}( V_A(\mathcal{\hat{F}},y)),
\end{equation}
where $\mathcal{\hat{F}}$ is the subset of the original feature set $\mathcal{F}$ and the generated new features $\mathcal{F}^g$ , so $\mathcal{\hat{F}} \subseteq  \mathcal{F} \bigcup \mathcal{F}^g$, where $\mathcal{F}^g$ is produced by applying $\mathcal{O}$ to $\mathcal{F}$.
\fi

%% file: method.tex
\section{Optimal and Explainable  Feature Space Reconstruction}
%In this section, we will introduce each technical component of our framework and its associated pseudo algorithm.
In this section, we present an overview, and then detail each technical component of our framework.

\subsection{Framework Overview}
Figure \ref{fig:framework} shows our proposed framework, \textbf{G}roup-wise \textbf{R}einforcement \textbf{F}eature \textbf{G}eneration (\textbf{GRFG}).
In the first step, we cluster the original features into different feature groups by maximizing intra-group feature cohesion and inter-group feature distinctness. 
In the second step, we leverage a group-operation-group strategy to cross two feature groups to generate multiple features each time. For this purpose, we develop a novel cascading reinforcement learning method to learn three agents to select the two most informative feature groups and the most appropriate operation from the operation set.
As a key enabling technique, the cascading reinforcement method will coordinate the three agents to share their perceived states in a cascading fashion, i.e., (agent1, state of the first feature group), (agent2, fused state of the operation and agent1's state), and (agent3, fused state of the second feature group and agent2's state), in order to learn better choice policies. 
After the two feature groups and operation are selected, we generate new features via a group-group crossing strategy. 
In particular, if the operation is unary, e.g., sqrt, we choose the feature group of higher relevance to target labels from the two feature groups, and apply the operation to the more relevant feature group  to generate new features. 
if the operation is binary,  we will choose the K most distinct feature-pairs from the two feature groups, and apply the binary operation to the chosen feature pairs to generate new features.
In the third step, we add the newly generated features to the original features to  create a generated feature set.
We feed the generated feature set into a downstream task to collect predictive accuracy as reward feedback to update  policy parameters.
Finally, we employ feature selection to eliminate redundant features and control the dimensonality of the newly generated feature set, which will be used as the original feature set to restart the iterations to regenerate new features until the maximum number of iterations is reached.

\subsection{Generation-oriented Feature Grouping}
One of our key findings is that group-wise feature generation can accelerate the generation and exploration, and, moreover, augment reward feedback of agents to learn clear policies. 
Inspired by this finding, our system starts with generation oriented feature clustering, which aims to create feature groups that are meaningful for group-group crossing.  
Our another insight is that crossing features of high (low) information distinctness is more (less) likely to generate meaningful variables in a new feature space. 
As a result, unlike classic clustering, the objective of feature grouping is to maximize inter-group feature information distinctness and minimize intra-group feature information distinctness.
To achieve this goal, we propose the \textbf{M-Clustering} for feature grouping, which starts with each feature as a feature group and then merges its most similar feature group pair at each iteration. 

\textbf{Distance Between Feature Groups: A Group-level Relevance-Redundancy Ratio Perspective.} To achieve the goal of minimizing intra-group feature distinctness and maximizing inter-group feature distinctness,  we develop a new distance measure to quantify the distance between two feature groups. We highlight two interesting findings: 1) relevance to predictive target: if the relevance between one feature group and predictive target is similar to the relevance of another feature group and predictive target,  the two feature groups are similar; 2) mutual information: if the mutual information between the features of the two groups are large, the two feature groups are similar. 
Based on the two insights, we devise a feature group-group distance. 
The distance can be used to evaluate the distinctness of two feature groups, and, further, understand how likely crossing the two feature groups will generate more meaningful features. Formally, the distance is given by:
\begin{equation}
\small
% \vspace{-0.18cm}
    \label{fea_dis}
    dis(\mathcal{C}_i, \mathcal{C}_j) =
    \frac{1}{|\mathcal{C}_i|\cdot|\mathcal{C}_j|}
    \sum_{f_i\in \mathcal{C}_i}\sum_{f_j\in \mathcal{C}_j}
    \frac{|MI(f_i,y)-MI(f_j,y)|}{MI(f_i,f_j)+\epsilon},
\end{equation}
where $\mathcal{C}_i$ and $\mathcal{C}_j$ denote two feature groups, $|\mathcal{C}_i|$ and $|\mathcal{C}_j|$  respectively are the numbers of features in $\mathcal{C}_i$ and $\mathcal{C}_j$, $f_i$ and $f_j$ are two features in $\mathcal{C}_i$ and $\mathcal{C}_j$ respectively, $y$ is the target label vector.
In particular, $|MI(f_i,y)-MI(f_j,y)|$ quantifies the  difference in relevance  between $y$ and $f_i$, $f_j$.
If $|MI(f_i,y)-MI(f_j,y)|$ is small,   $f_i$ and $f_j$ have a more similar influence on classifying $y$;
$MI(f_i,f_j)+\epsilon$ quantifies the redundancy between $f_i$ and $f_j$.
$\epsilon$ is a small value that is used to prevent the denominator from being zero.
If $MI(f_i,f_j)+\epsilon$ is big, $f_i$ and $f_j$ share more information.
 
\textbf{Feature Group Distance based M-Clustering Algorithm:}
We develop a group-group distance instead of point-point distance,  and under such a group-level distance, the shape of the feature cluster could be non-spherical. Therefore, it is not appropriate to use K-means or density based methods. Inspired by agglomerative clustering,  given a feature set $\mathcal{F}$, we propose a three step method: 1) INITIALIZATION:  we regard each feature in $\mathcal{F}$ as a small feature cluster. 
2) REPEAT:  we calculate the information overlap between any two feature clusters and determine which cluster pair is the most closely overlapped.  We then merge two clusters into one and remove the two original clusters.
3) STOP CRITERIA:  we iterate the REPEAT step until the distance between the closest feature group pair reaches a certain threshold. Although classic stop criteria is to stop when there is only one cluster, using the distance between the closest feature groups as stop criteria can better help us to semantically understand, gauge, and identify the information distinctness among feature groups. It eases the implementation in practical deployment. 
% \vspace{-0.5cm}

%Before introducing our measurement, we firstly provide the definition of MI~\cite{cover1999elements}. The MI between two discrete variable vectors is: $%\begin{equation}\label{mutual_info} MI(X_1,X_2) = \sum_{x_1\in X_1}\sum_{x_2\in X_2} p(x_1,x_2)log(\frac{p(x_1,x_2)}{p(x_1)p(x_2)}), %\end{equation} $where $X_1$ and $X_2$ are two vectors, $x_1$ and $x_2$ are elements of $X_1$ and $X_2$ respectively, $p(x_1,x_2)$ is the joint distribution of $X_1$ and $X_2$, $p(x_1)$ and $p(x_2)$ are the marginal distribution of $X_1$ and $X_2$ respectively. The greater the value of MI, the more information the two vectors share. By using this distance metric, we can group the features that share similar information and have similar impact on identifying target label into a single feature cluster. Our framework begins by clustering the feature set into  feature clusters. This step groups similar features into a single cluster and dissimilar features into different clusters. This is because we presume that the generated features are meaningful when the interacted features come from different information distributions; otherwise, the generated features should be redundant. Considering that the boundary and information distribution of feature clusters vary over the feature generation process, we develop a new  M-Clustering algorithm rather than using a conventional clustering technique.

\subsection{Cascading Reinforcement Feature Groups and Operation Selection}
\begin{figure}[!t]
    \centering
    \includegraphics[width=0.65\linewidth]{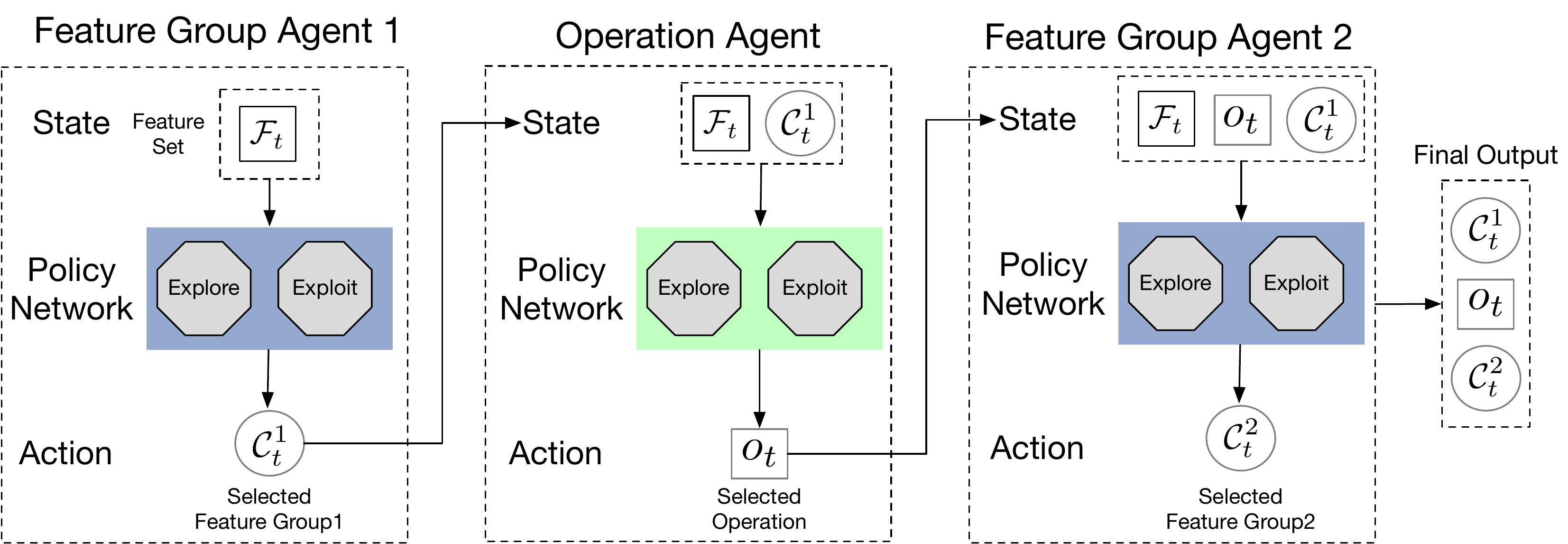}
    % \vspace{-0.2cm}
    \captionsetup{justification=centering}
    \vspace{-0.1cm}
    \caption{The cascading agents are comprised of the feature group agent1, the operation agent, and the feature group agent2. They collaborate to choose two candidate feature groups and a single operation.}
    \vspace{-0.4cm}
    \label{fig:agents}
\end{figure}

To achieve group-wise feature generation, we need to  select a feature group, an operation, and another feature group to perform group-operation-group based crossing. Two key findings inspire us to leverage cascading reinforcement. 
\textbf{Firstly}, we highlight that although it is hard to define and program the optimal selection criteria of feature groups and operation, we can view selection criteria as a form of machine-learnable policies. 
We propose three agents to learn the policies by trials and errors. 
\textbf{Secondly}, we find that the three selection agents are cascading in a sequential auto-correlated decision structure, not independent and separated. 
Here, ``cascading'' refers to the fact that within each iteration agents  make decision sequentially, and downstream agents await for the completion of an upstream agent.  The decision of an upstream agent will change the environment states of downstream agents.
As shown in Figure ~\ref{fig:agents}, the first agent picks the first feature group based on the state of the entire feature space, the second agent picks the operation based on the entire feature space and the selection of the first agent, and the third agent chooses the second feature group based on the entire feature space and the selections of the first and second agents. 

We next propose two generic metrics to quantify the usefulness (reward) of  a feature set, and then form three MDPs to learn three selection policies.

%MDP is a discrete-time stochastic control process, which can be defined as a 4-tuple $\mathcal{M}=\{\mathcal{S},\mathcal{A},\mathcal{P}, \mathcal{R}\}$. In this tuple, $\mathcal{S}$ denotes the state space, $\mathcal{A}$ denotes the action space, $\mathcal{P}$ indicates the transition probability that the current action in the current state will lead the next state. $\mathcal{R}$ indicates the immediate reward gained as a result of the state transition through taking action.

\smallskip
\noindent\textbf{Two Utility Metrics for Reward Quantification.} The two utility metrics are from  the supervised and unsupervised perspectives. 

\noindent\ul{Metric 1: Integrated Feature Set Redundancy and Relevance.}  
We propose a metric to quantify feature set utility from an information theory perspective: a higher-utility feature set has less redundant information and is more relevant to prediction targets.
Formally, given a feature set $\mathcal{F}$ and a predictive target label $y$, such utility metric can be calculated by 
\begin{equation}
\small
    U(\mathcal{F}|y) = -\frac{1}{|\mathcal{F}|^2}\sum_{f_i, f_j \in \mathcal{F}} MI(f_i, f_j) + \frac{1}{|\mathcal{F}|}\sum_{f\in \mathcal{F}}MI(f,y),
\end{equation}
where $MI$ is the mutual information, $f_i, f_j, f$ are features in $\mathcal{F}$ and $|\mathcal{F}|$ is the size of the feature set $\mathcal{F}$.

\noindent\ul{Metric 2: Downstream Task Performance.} 
Another utility metric is whether this feature set can improve a downstream task (e.g., regression, classification). We use a downstream predictive task performance indicator (e.g., 1-RAE, Precision, Recall, F1) as a utility metric of a feature set. 

%We assess the utility of the feature set $\mathcal{F}$ by performing a downstream task (\textit{e.g.} regression or classification). The value of the quantitative indicator  (\textit{e.g.} 1-relative absolute error or f1) of the task (denoted by $V_A$) is the reflection of the utility of $\mathcal{F}$ from another perspective. Intuitively, a high $V_A$ value suggests that the $\mathcal{F}$ has an excellent distinguishable ability to the target label $y$.

\smallskip
\noindent\textbf{Learning Selection Agents of Feature Group 1, Operation, and Feature Group 2.} 
Leveraging the two metrics, we next develop three MDPs to learn three agent policies to select the best feature group 1, operation, feature group 2. 

\noindent\ul{\textit{Learning the Selection Agent of Feature Group 1.}} 
The feature group agent 1 iteratively select the best meta feature group 1.
Its learning system includes: \textbf{i) Action:} its action 
at the $t$-th iteration is the meta feature group 1  selected from the feature groups of the previous iteration, denoted group $a_t^1 = \mathcal{C}^1_{t-1}$.
\textbf{ii) State:} its state at the $t$-th iteration is a vectorized representation of the generated feature set of the previous iteration.
Let $Rep$ be a state representation method, the state can be denoted by $s_t^1 = Rep(\mathcal{F}_{t-1})$.
We will discuss the state representation method in the Section~\ref{rep_fun}.
\textbf{iii) Reward:} its reward at the $t$-th iteration is the utility score the selected feature group 1, denoted by  $\mathcal{R}(s_t^1,a_t^1)=U(\mathcal{F}_{t-1}|y)$.
% Formally, let $Rep$ be a state representation method, in the $t$-th iteration, after the feature group agent 1 takes an action $a_t^{1}$ to select the meta feature group $\mathcal{C}^1_t$ under the state of the previous iteration's feature set $s_t^1 = Rep(\mathcal{F}_{t-1})$. We provide the reward $U(\mathcal{F}_{t-1}|y)$ to the agent.

%can be defined by $\mathcal{M}^1 =\{\mathcal{S}^1,\mathcal{A}^1,\mathcal{P}^1, \mathcal{R}^1\}.$ The key components are as follows: (1) \textbf{State.}  The $t$-th state in $\mathcal{S}^1$ is the representation of the feature set $\mathcal{F}_t$ at the $t$-th iteration, denoted by $s^1_t = \phi(\mathcal{F}_t)$. The representing process $\phi$ will be illustrated in the following section. (2) \textbf{Action.} The $t$-th action is the selected feature cluster1, denoted by  $a^1_t = \mathcal{C}^1_t$. (3) \textbf{Transition.} Owing to $\mathcal{M}^1$ is deterministic, the transition probability from $s^1_t$ to $s^1_{t+1}$ via action $a^1_t$ is 1, denoted by $\mathcal{P}^1(s^1_t,a^1_t,s^1_{t+1})=1$.

\noindent\ul{\textit{Learning the Selection Agent of Operation.}} 
The operation agent will iteratively select the best operation (\textit{e.g.} +, -) from an operation set as a feature crossing tool for feature generation.
Its learning system includes: \textbf{i) Action:} its action at the $t$-th iteration is the selected operation, denoted by $a_t^o = o_t$.
\textbf{ii) State:} its state at the $t$-th iteration is the combination of $Rep(\mathcal{F}_{t-1})$ and the representation of the  feature group selected by the previous agent, denoted by $s^o_t = Rep(\mathcal{F}_{t-1}) \oplus Rep(\mathcal{C}_{t-1}^1)$, where $\oplus$ indicates the concatenation operation. 
\textbf{iii) Reward:} The selected operation will be used to generate new features by feature crossing. We combine such new features with the original feature set to form a new feature set.
Thus, the feature set at the $t$-th iteration is $\mathcal{F}_t = \mathcal{F}_{t-1} \cup g_t$, where $g_t$ is the generated new features.
The reward of this iteration is the improvement in the utility score of the feature set compared with the previous iteration, denoted by $\mathcal{R}(s_t^o,a_t^o) = U(\mathcal{F}_t|y) - U(\mathcal{F}_{t-1}|y)$.

%\textbf{MDP for Operator Selection} can be formulated by $\mathcal{M}^o =\{\mathcal{S}^o,\mathcal{A}^o,\mathcal{P}^o, \mathcal{R}^o\}.$  The key components are as follows: (1) \textbf{State.} The $t$-th state in $\mathcal{S}^{o}$ is the combination of $\phi(\mathcal{F}_t)$ and the representation of  the selected feature cluster $\mathcal{C}_t^1$ in $\mathcal{M}^1$ at the $t$-th iteration, denoted by $s^o_t =  concat(\phi(\mathcal{F}_t),  \phi(\mathcal{C}^1_t)$). (2) \textbf{Action.} The $t$-th action is the selected operator from the operator set, denoted by $a^o_t=o_t$. (3) \textbf{Transition.} The same to $\mathcal{M}_1$, the transition probability from $s^o_t$ to $s^o_{t+1}$ via action $a^o_t$ is 1, denoted by $\mathcal{P}^o(s^o_t,a^o_t,s^o_{t+1})=1$.

\noindent\ul{\textit{Learning the Selection Agent of Feature Group 2.}} 
The feature group agent 2 will iteratively select the best meta feature group 2 for feature generation.
Its learning system includes: \textbf{i) Action:} its action at the $t$-th iteration is the meta feature group 2 selected from the clustered feature groups of the previous iteration, denoted by $a_t^2 = \mathcal{C}^2_{t-1}$.
\textbf{ii) State:} its state at the $t$-th iteration is combination of $Rep(\mathcal{F}_{t-1})$, $Rep(\mathcal{C}_{t-1}^1)$ and the vectorized representation of the  operation selected by the operation agent, denoted by $s_t^2 = Rep(\mathcal{F}_{t-1})\oplus Rep(\mathcal{C}_{t-1}^1) \oplus Rep(o_t)$.
\textbf{iii) Reward:} its reward at the $t$-th iteration is improvement of the feature set utility and the feedback of the downstream task, denoted by $\mathcal{R}(s_t^2, a_t^2) = U(\mathcal{F}_t|y)-U(\mathcal{F}_{t-1}|y)+V_{A_t}$, where $V_A$ is the performance (e.g., F1) of a downstream predictive task. 

% \begin{figure}[t]
%     \centering
%     \includegraphics[width=1.0\linewidth]{img/state_ds.pdf}
%     % \vspace{-0.3cm}
%     \captionsetup{justification=centering}
%     % \vspace{-0.2cm}
%     \caption{\color{red} The illustration of Descriptive Statistics Based State Method.}
%     % \vspace{-0.3cm}
%     \label{fig:state_repr_ds}
% \end{figure}

% \begin{figure}[t]
%     \centering
%     \includegraphics[width=1.0\linewidth]{img/state_ae.pdf}
%     % \vspace{-0.3cm}
%     \captionsetup{justification=centering}
%     % \vspace{-0.2cm}
%     \caption{\color{blue} The illustration of Autoencoder Based State Method.}
%     % \vspace{-0.3cm}
%     \label{fig:state_repr_ae}
% \end{figure}

% \begin{figure}[t]
%     \centering
%     \includegraphics[width=1.0\linewidth]{img/state_cg.pdf}
%     % \vspace{-0.3cm}
%     \captionsetup{justification=centering}
%     % \vspace{-0.2cm}
%     \caption{\color{blue} The illustration of Correlation Graph Based State Method.}
%     % \vspace{-0.3cm}
%     \label{fig:state_repr_cg}
% \end{figure}

\begin{figure*}[t]
    \centering
    \vspace{-0.3cm}
    \includegraphics[width=1.0\linewidth]{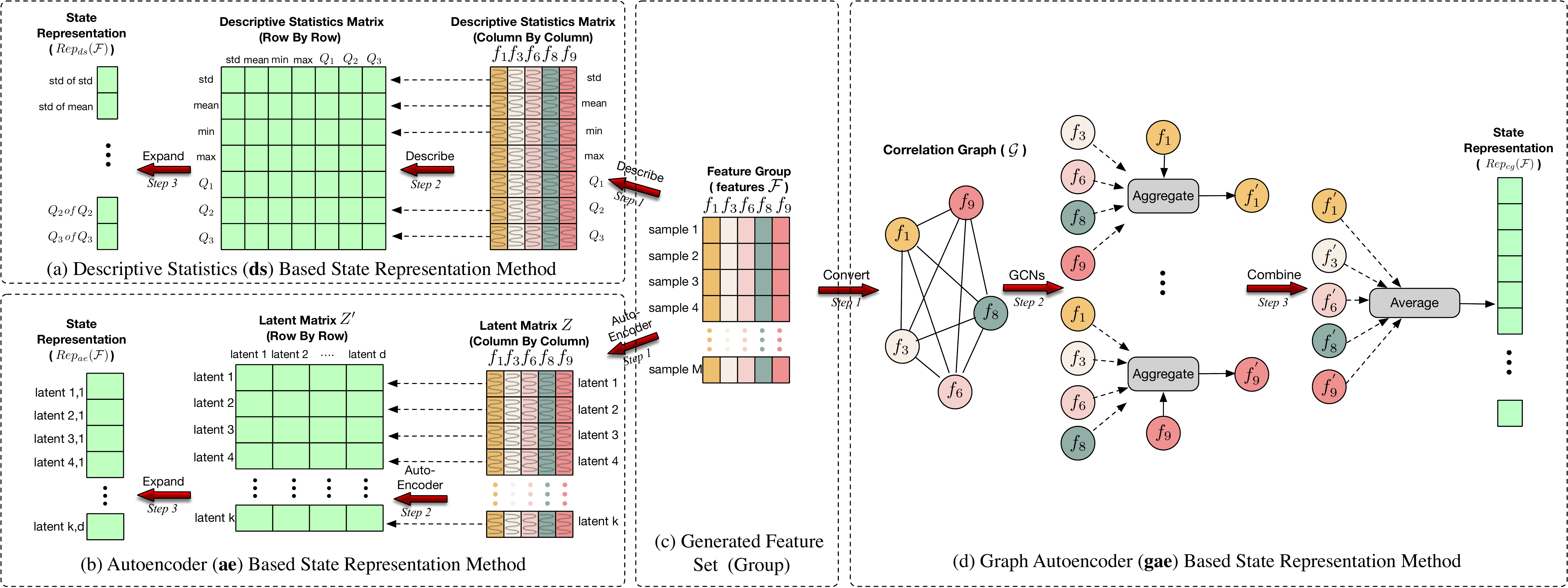}
    \captionsetup{justification=centering}
    \vspace{-0.3cm}
    \caption{Three state representation methods for a feature set (group): (a) Descriptive statistics based state representation  \textbf{ds}; (b) Autoencoder based state representation  \textbf{ae}; (c) Graph autoencoder based state representation  \textbf{gae}.}
    \vspace{-0.5cm}
    \label{fig:state_repr}
\end{figure*}

% 把这部分按dueling double dqn的形式改写这章， 把公式写过来
% \smallskip
\subsection{Optimization Technique}~\label{ot}
In the preliminary work~\cite{wang2022multi}, we use the value-based optimization technique to train the entire framework. 
To further enhance the generalization ability and performance of our framework, we use a brand new optimization perspective, actor-critic, to learn the feature transformation policies.

% {\color{blue} In the previous section, we illustrate the framework of the Group-wise Reinforcement Feature Generation, and in conference version, we use the basic idea decrived from Q-learning to optimize the whole process. In this journal version, we thoroughly examine two dimensions of reinforcement learning optimization techniques, thereby expanding the potential of the framework.}

\smallskip
\noindent\textbf{Value-based Optimization Perspective.} 
We train the three agents by maximizing the discounted and cumulative reward during the iterative feature generation process.
In other words, we encourage the cascading agents to collaborate to generate a feature set that is independent, informative, and performs well in the downstream task. For each agent, we firstly give the memory unit of the $t$-th exploration as $\mathcal{M}_t = (s_t, a_t, r_t, s_{t+1}, a_{t+1})$.
In the preliminary work, we minimize the temporal difference error $\mathcal{L}$ based on Q-learning.
The optimization objective can be formulated as:
\begin{equation}
\label{Q_update}
    \mathcal{L} = Q(s_{t},a_t) - (\mathcal{R}_t + \gamma * \text{max}_{a_{t+1}}Q(s_{t+1},a_{t+1})),
\end{equation}
where $Q(\cdot)$ is the Q function
that estimates the value of choosing action $a$ based on the state $s$.
$\gamma$ is a discount value, which is a constant between 0 and 1.
It raises the significance of rewards in the near future and diminishes the significance of rewards in the uncertain distant future, ensuring the convergence of rewards.

However, Q-learning frequently overestimates the Q-value of each action, resulting in suboptimal policy learning.
The reason for this is that traditional Q-learning employs a single Q-network for both Q-value estimation and action selection.
The max operation in Equation~\ref{Q_update} will accumulate more optimistic Q-values during the learning process, leading to Q-value overestimation.

\smallskip
\noindent\textbf{Improving the Q-learning.}
In this journal version, we propose a new training strategy to alleviate this issue, namely, GRFG$^\upsilon$.
Specifically, we use two independent Q-networks with the same structure to conduct Q-learning.
The first Q-network is used to select the action with the highest Q-value.
The index of the selected action is used to query the Q-value in the second Q-network for Q-value estimation.
The optimization goal can be formulated as follows: 
\begin{equation}
\label{double_Q_update}
    \mathcal{L}^\upsilon = Q(s_{t},a_t) - (r_t + \gamma * \text{max}_{a_{t+1}}Q^{-}(s_{t+1},a_{t+1})),
\end{equation}
where $Q(\cdot)$ and $Q^{-}(\cdot)$ are two independent networks with the same structure. 
The parameters of them are not updated simultaneously. 
The asynchronous learning strategy reduces the possibility that the maximum Q-value of the two networks is associated with the same action, thereby alleviating Q-value overestimation.

Furthermore, to make the learning of each Q-network robust, we decouple the Q-learning into: state value estimation and action advantage value estimation~\cite{wang2016dueling}.  
Thus, the calculation of Q-value can be formulated as follows:
\begin{equation}
\label{q}
    Q_\upsilon(s, a) = V_\upsilon(s) + A_\upsilon(s, a),
\end{equation}
where $V_\upsilon(\cdot)$ and $A_\upsilon(\cdot)$ are state value function and action advantage value function respectively.
This decoupling technique generalizes action-level learning without excessive change, resulting in a more robust and consistent Q-learning.

After agents converge, we expect to find the optimal policy $\pi^*_\upsilon$ that can choose the most appropriate action (\textit{i.e.} feature group or operation) based on the state via the Q-value, which can be formulated as follows:
\begin{equation}
    \label{rl_policy}
    \pi^*_\upsilon(a_t|s_t) = \argmax_{a} Q_\upsilon(s_t,a).
\end{equation}
\smallskip
% \vspace{-0.2cm}

\smallskip
\noindent\textbf{Actor-Critic Optimization Perspective.} 
We adopt the Actor-Critic optimization approach~\cite{haarnoja2018soft, lillicrap2015continuous} to enhance the convergence and performance of the model, namely, GRFG$^\alpha$. Actor-Critic architecture consists of two independent network, which is the \textit{Actor} network for action selection and \textit{Critic} network for reward estimation. 

\noindent\textit{\ul{Actor}}: The goal of the actor is to learn the selection policy, denoted by $\pi_\alpha(\cdot)$, based on the current state. This policy is used to select appropriate candidate feature groups or operations. In the $t$-th iteration, given the state $s_t$, the agent chooses an action $a_t$ as follows:
\begin{equation}
\label{actor}
a_t \sim \pi_\alpha(s_{t}),
\end{equation}
where the output of $\pi_\alpha(\cdot)$ is the probability distribution over candidate actions. The $\sim$ operation denotes the sampling operation.

\noindent\textit{\ul{Critic}}: The critic's objective is to estimate the expected reward of a given input state, which can be expressed as:
\begin{equation}
v_t = V_\alpha(s_t),
\end{equation}
In this equation, $V_\alpha(\cdot)$ is the state-value function, and $v_t$ represents the resulting value. 

\noindent\textit{\ul{Optmization}}: 
We update the \textit{Actor} and \textit{Critic} in cascading agents after each step of feature transformation.  
Suppose in step-$t$, for each agent, we can obtain the memory as $\mathcal{M}_t = (a_t, s_t, s_{t+1}, r_t)$. 
The estimation of temporal difference error ($\delta$) for Critic is given by:
\begin{equation}
  \label{advantage}
    \delta = r_t + \gamma V_\alpha(s_{t+1}) - V_\alpha(s_t),
\end{equation}
where $\gamma \in [0, 1]$ is the discounted factor. By that, we can update the Critic by mean square error (MSE):
\begin{equation}
    \mathcal{L}^\alpha_c = \delta^2
\end{equation}
Then, we can update the policy $\pi_\alpha$ in Actor by policy gradient, defined as:
\begin{equation}
    \mathcal{L}_a^\alpha = log\pi_\alpha (a_t|s_t) * \delta + \beta * H(\pi_\alpha (s_t)),
\end{equation}
where $H(\cdot)$ is an entropy regularization term that aims to increase the randomness in exploration. We use $\beta$ to control the strength of the $H$. $\pi_\alpha (a_t|s_t)$ denote the probability of action $a_t$.
The overall loss function for each agent is given by:
\begin{equation}
    \mathcal{L}^\alpha = \mathcal{L}^\alpha_c + \mathcal{L}^\alpha_a.
\end{equation}
After agents converge, we expect to discover the optimal policy $\pi_\alpha^*$ that can choose the most appropriate action (\textit{i.e.} feature group or operation).

\subsection{State Representation of a Feature Set (Group) and an Operation.}\label{rep_fun}
State representation reflects the status of current features or operations, which serves as the input for reinforced agents to learn more effective policies in the future.
We propose to map a feature group or an operation to a vectorized representation  for agent training. 

For a feature set (group), we propose three state representation methods.
To simplify the description, in the following parts, suppose given the feature set $\mathcal{F}\in \mathbb{R}^{m\times n}$, where the $m$ is the number of the total samples, and the $n$ is the number of feature columns.

\noindent\textbf{Descriptive Statistics Based State Representation.}
In the preliminary work~\cite{10.1145/3534678.3539278}, we use the descriptive statistic matrix of a feature set as its state representation.
Figure~\ref{fig:state_repr} (a) shows its calculation process:
\noindent\textit{Step-1}: We calculate the descriptive statistics matrix (\textit{i.e.} count, standard deviation, minimum, maximum, first, second, and third quartile) of $\mathcal{F}$ column by column. 
\noindent\textit{Step-2}: We calculate the descriptive statistics of the outcome matrix of the previous step row by row to obtain the meta descriptive statistics matrix that shape is $\mathbb{R}^{7\times 7}$.
\noindent\textit{Step-3}: We obtain the feature-feature's state representation $Rep_{ds}(\mathcal{F})\in \mathbb{R}^{1\times 49}$ by flatting the descriptive matrix from the former step. 

While descriptive statistics based state representation has shown exceptional performance, it may cause information loss due to fixed statistics views.
To better comprehend the feature set, we propose two new state representation methods in the journal version work.

% 把loss function写过来，把过程部分的流程按公式写过来
\noindent\textbf{Autoencoder Based State Representation.}
Intuitively, an effective state representation can recover the original feature set.
Thus, we user an autoencoder  to convert  the feature set $\mathcal{F}$ into a latent embedding vector and regard it as the state representation.
Figure~\ref{fig:state_repr} (b) shows the calculation process, which can be devided to three steps:
Specifically, 
\noindent\textit{Step-1}: We apply a fully connected network $FC_{col}(\cdot)$ to convert each column of $\mathcal{F}$ into matrix $Z \in \mathbb{R}^{k\times n}$, where $k$ is the dimension of the latent representation of each column.
The process can be represented as:
\begin{equation}
    Z = FC_{col}(\mathcal{F}).
\end{equation}
\noindent\textit{Step-2}: We apply another fully connected network $FC_{row}(\cdot)$ to convert each row of $Z$ into matrix $Z' \in \mathbb{R}^{k\times d}$, where $d$ is the dimension of the latent embedding of each row.
The process can be defined as:
\begin{equation}
    Z' = FC_{row}(Z).
\end{equation}
\noindent\textit{Step-3}: We flatten $Z'$ and input it into a decoder constructed by fully connected layers to reconstruct the original feature space, which can be defined as:
\begin{equation}
\label{ae_emb}
\mathcal{F}' = FC_{decoder}(\text{Flatten}(Z')).
\end{equation}
During the learning process, we minimize the reconstructed error between $\mathcal{F}$ and $\mathcal{F}'$.
The optimization objective is as follows:
\begin{equation}
\mathcal{L}_{ae} = \text{MSELoss}(\mathcal{F}, \mathcal{F}').
\end{equation}
When the model converges,  $\text{Flatten}(Z')$ used in Equation~\ref{ae_emb} is the state representation $Rep_{ae}(\mathcal{F})$.
% We conducted experiments and noticed that the autoencoder would converge between 15 to 20 epochs. So, for every generated feature set (group), we train the autoencoder based state representation method for 20 epochs.

% 把loss function写过来，把过程部分的流程按公式写过来 （改写成Graph Autoencoder的形式）
\noindent\textbf{Graph Autoencoder Based State Representation.}
The prior two methods only preserve the information of individual features in the state representation while disregarding their correlations.
To further enhance state representation, we propose a graph autoencoder~\cite{kipf2016semi} to incorporate the knowledge of the entire feature set and feature-feature correlations into the state representation.
Figure~\ref{fig:state_repr} (d) shows the calculation process. Specifically,
\noindent\textit{Step-1}: We build a complete correlation graph $\mathcal{G}$ by calculating the similarity matrix between each pair of feature columns.
The adjacency matrix of $\mathcal{G}$ is $\mathbf{A} \in \mathbb{R}^{n\times n}$, where a node is a feature column in $\mathcal{F}$ and an edge reflects the similarity between two nodes. 
Then, we use the encoder constructed by one-layer GCN to aggregate feature knowledge to generate  enhanced embeddings $Z\in \mathbb{R}^{n\times k}$ based on $\mathbf{A}$, where $k$ is the dimension of latent embedding. The calculation process is defined as follows:
\begin{equation}
    Z = ReLU(\mathbf{D}^{-\frac{1}{2}}\mathbf{A}\mathbf{D}^{-\frac{1}{2}}\mathcal{F}^{\top}\mathbf{W}),
\end{equation}
where $\mathcal{F}^{
\top}$ is the transpose of input feature $\mathcal{F}$, $\mathbf{D}$ is the degree matrix, and  $\mathbf{W} \in \mathbb{R}^{m\times k}$ is the parameter of GCN. 
\noindent\textit{Step-2}:
Then, we use the decoder to reconstruct the adjacency matrix $\mathbf{A}'$ based on  $Z$, which can be represented as:
\begin{equation}
   \mathbf{A}' = \text{Sigmoid}(Z\cdot Z^{\top}).
\end{equation}
During the learning process, we minimize the reconstructed error between $\mathbf{A}$ and $\mathbf{A}'$.
The optimization objective is:
\begin{equation}
\mathcal{L}_{gae} = \text{BCELoss}(\mathbf{A}, \mathbf{A}').
\end{equation}
where BCELoss refers to the binary cross entropy loss function.
When the model converges, we average $Z$ column-wisely to get the state representation $Rep_{gae}(\mathcal{F}) \in \mathbb{R}^{1\times k}$.
For an operation $o$, we use its one hot vector as the sate representation, which can be denoted as $Rep(o)$.

\subsection{Group-wise Feature Generation}
\label{group_generation}
We found that using group-level crossing can generate more features each time, and, thus, accelerate exploration speed, augment reward feedback by adding significant amount of features, and effectively learn policies.  
The selection results of our reinforcement learning system include \textbf{two generation scenarios}: (1) selected are a binary operation and two feature groups;  (2) selected are a unary operation and two feature groups. 
However, a challenge arises: what are the most effective generation strategy for the two scenarios?
We next propose two  strategies for the two scenarios.

\noindent\textbf{Scenario 1: Cosine Similarity Prioritized Feature-Feature Crossing.} 
We highlight that it is more likely to generate informative features by crossing two features that are less overlapped in terms of information. 
We propose a simple yet efficient strategy, that is, to select the top K dissimilar feature pairs between two feature groups. 
Specifically, we first cross two selected feature groups to prepare feature pairs. We then compute the cosine similarities of all feature pairs. Later, we rank and select the top K dissimilar feature pairs. Finally, we apply the operation to the top K selected feature pairs to generate K new features.

\noindent\textbf{Scenario 2: Relevance Prioritized  Unary Feature Generation. }
When selected  are an unary operation and two feature groups, we directly apply the operation to the feature group that is more relevant to target labels. Given a feature group $C$, we use the average mutual information between all the features in $C$ and the prediction target $y$ to quantify the relevance between the feature group and the prediction targets, which is given by: 
$
   rel =  \frac{1}{|\mathcal{C}|}\sum_{f_i\in \mathcal{C}} MI(f_i,y) 
$, 
where $MI$ is a function of mutual information. 
After the more relevant feature group is identified, we apply the unary operation to the feature group to generate new ones. 

\iffalse
\textbf{Mutual Information based Feature Cluster Selection.} When the operator is unary, we employ this interaction manner. To generate more significant features, we attempt to chose the feature cluster that is more closely related to the target label. Specifically, for two feature clusters $\mathcal{C}^1_t, \mathcal{C}^2_t$, we  uses the average relevance between $\mathcal{C}^1_t$ and $y$ to subtract the average relevance between $\mathcal{C}^2_t$ and $y$ to calculate the difference $\delta$. The process can be formulated as:
\begin{equation}
    \label{differ}
    \delta = ( \frac{1}{|\mathcal{C}^1_t|}\sum_{f^1_i\in \mathcal{C}^1_t} MI(f^1_i,y) -  \frac{1}{|\mathcal{C}^2_t|}\sum_{f^2_j\in \mathcal{C}^2_t} MI(f^2_j,y)),
\end{equation}
where $|\mathcal{C}^1_t|, |\mathcal{C}^2_t|$ are the number of features in $\mathcal{C}^1_t, \mathcal{C}^2_t$ respectively; 
$f_i^1, f_j^2$ are the single feature in $\mathcal{C}^1_t, \mathcal{C}^2_t$ respectively. if $\delta \geq 0$, which means $\mathcal{C}_t^1$ is more relevant to $y$ than $\mathcal{C}_t^2$.  Thus, we apply the selected operator $o_t$ to $\mathcal{C}^1_t$ to generate new features, which is defined as $\mathcal{F}_t^g = o_t(\mathcal{C}^1_t)$.
Otherwise, $\mathcal{C}_2^t$ is our selection, we generate new features by applying $o_t$ to $\mathcal{C}^2_t$, which is formulated as $\mathcal{F}_t^g = o_t(\mathcal{C}^2_t)$.
\fi

\noindent\textbf{Post-generation Processing.} 
After feature generation, we combine the newly generated features with  the original feature set to form an updated feature set, which will be fed into a downstream task to evaluate predictive performance.
Such performance is exploited as reward feedback to update the policies of the three cascading agents in order to optimize the next round of feature generation.
To prevent feature number explosion during the iterative generation process, 
we use a feature selection step to control feature size. 
When the size of the new feature set exceeds a feature size tolerance threshold, we leverage the K-best feature selection method to reduce the feature size. Otherwise, we don't perform feature selection. 
We use the tailored new feature set as the original feature set of the next iteration. 

Finally, when the maximum number of iterations is reached, the algorithm returns the optimal feature set $\mathcal{F^{*}}$ that has the best performance on the downstream task over the entire exploration.

\subsection{Applications}
Our proposed method (GRFG) aims to automatically optimize and generate a traceable and optimal feature space for an ML task.
This framework is applicable to a variety of domain tasks without the need for prior knowledge.
We apply GRFG to classification, regression, and outlier detection tasks on 29 different tabular datasets to validate its performance.

%% file: experiment.tex
\section{Experiments}
%We detail experimental setup in detail, and then present experimental results.
 
% \iffalse
% We design experiments to answer the following research questions: \textbf{Q1.} Is the performance of our feature generation framework (GRFG) superior to existing baseline models? 
% \textbf{Q2.} {\color{red} How is the learning process of GRFG interpreted?}
% \textbf{Q3.} How does each component in GRRG impact its performance?
% \textbf{Q4.} Is M-clustering more effective than traditional clustering algorithms in improving feature generation?
% \textbf{Q5.} Is GRFG robust with different machine learning models as downstream tasks?
% \textbf{Q6.} Is the feature generation process of GRFG traceable and explainable?
% \fi

\subsection{Experimental Setup}
\subsubsection{Data Description}
We used 29 publicly available datasets from UCI~\cite{uci}, 
 LibSVM~\cite{libsvm},
 Kaggle~\cite{kaggle}, and OpenML~\cite{openml} to conduct experiments.
The 24 datasets involves 14 classification tasks, 10 regression tasks, and 5 outlier detection tasks.
Table \ref{table_overall_perf} shows the statistics of the data. 

% % \vspace{-0.1cm}
\subsubsection{Evaluation Metrics}
We used the F1-score to evaluate the recall and precision of classification tasks. 
We used  1-relative absolute error (RAE) to evaluate the accuracy of regression tasks. 
Specifically, $\text{1-RAE} = 1 - \frac{\sum_{i=1}^{n} |y_i-\check{y}_i|}{\sum_{i=1}^{n}|y_i-\bar{y}_i|}$, where $n$ is the number of data points, $y_i, \check{y}_i, \bar{y}_i$ respectively denote golden standard  target values, predicted target values, and the mean of golden standard targets. 
For the outlier detection task, we adopted the Area Under the Receiver Operating Characteristic Curve (ROC/AUC) as the metric. 
To better evaluate the model performance among each dataset, we statistic the ranking of each model on every dataset and average this rank to calculate the Average Ranking.
%Whether in classification or regression, the higher the value of the evaluation metric, the more accurate the model.

\subsubsection{Baseline Algorithms}
\label{baseline}
We compared our method with five widely-used feature generation methods: 
(1) \textbf{RDG} randomly selects feature-operation-feature pairs for feature generation; 
(2) \textbf{ERG} is a expansion-reduction method, that applies operations to each feature to expand the feature space and selects significant features from the larger space as new features. 
(3) \textbf{LDA}~\cite{blei2003latent} extracts latent features via matrix factorization.
(4) \textbf{AFT}~\cite{horn2019autofeat}  is an enhanced ERG implementation that iteratively explores feature space and adopts a multi-step feature selection relying on L1-regularized linear models.
(5) \textbf{NFS}~\cite{chen2019neural} mimics feature transformation trajectory for each feature and optimizes the entire feature generation process through reinforcement learning.
(6) \textbf{TTG}~\cite{khurana2018feature} records the feature generation process using a transformation graph, then uses reinforcement learning to explore the graph to determine the best feature set.

Besides, we developed four variants of GRFG to validate the impact of each technical component: 
(i) $\textbf{GRFG}^{-c}$ removes the clustering step of GRFG and generate features by feature-operation-feature based crossing, instead of group-operation-group based crossing. 
(ii) $\textbf{GRFG}^{-d}$ utilizes the euclidean distance as the measurement of M-Clustering.
(iii) $\textbf{GRFG}^{-u}$ randomly selects a feature group from the feature group set, when the operation is unary.
(iv) $\textbf{GRFG}^{-b}$ randomly selects features from the larger feature group to align two feature groups when the operation is binary. 
We adopted Random Forest for classification  and regression tasks and applied K-Nearest Neighbors for outlier detection, in order to ensure the changes in results are mainly caused by the feature space reconstruction, not randomness or variance of the predictor.
We performed 5-fold stratified cross-validation in all  experiments.
Additionally, to investigate the impacts of the optimization strategies, we developed two model variants of GRFG: GRFG$^\upsilon$ (adopted the \textit{Improving Q-learning strategy}) and GRFG$^\alpha$ (adopted the \textit{Actor-Critic optimization perspective}).

\subsubsection{Hyperparameters, Source Code and Reproducibility}
The operation set consists of \textit{square root, square, cosine, sine, tangent, exp, cube, log, reciprocal, sigmoid, plus, subtract, multiply, divide}.
We limited iterations (epochs) to 30, with each iteration consisting of 15 exploration steps.
When the number of generated features is twice of the original feature set size, we performed feature selection to control feature size.
In GRFG, all agents were constructed using a DQN-like network with two linear layers activated by the RELU function.
We optimized DQN and its model variants using the Adam optimizer with a 0.01 learning rate, and set the limit of the experience replay memory as 32 and the batch size as 8.
The training epochs for autoencoder and graph autoencoder-based state representation have been set to 20 respectively.
The parameters of all baseline models are set based on the default settings of corresponding papers.
% For other detailed experimental settings, please check the code released in the Abstract section.

\subsubsection{Environmental Settings}
All experiments were conducted on the Ubuntu 18.04.6 LTS operating system, AMD EPYC 7742 CPU and 8 NVIDIA A100 GPUs, with the framework of Python 3.9.10 and PyTorch 1.8.1.

% \vspace{-0.1cm}
\subsection{Experimental Results}

\begin{table*}[!htbp]
\centering
\vspace{-0.1cm}
\caption{Overall performance comparison. `C' for classification, `R' for regression, and `D' for Outlier Detection. For robustness check, we report the average performance ranking of each method on all datasets in the last row (less is better).}
\vspace{-0.3cm}
\label{table_overall_perf}
\setlength{\tabcolsep}{1mm}{
% {|l|l|l|l|l|l|l|l|l|l|l|l|}{c|c|c|c|c|c|c|c|c|c|c|c}
\resizebox{\linewidth}{!}{
\begin{tabular}{cccccccccccccc}
\toprule
Dataset            & Source   & Task & Samples & Features & RDG  & {ERG} & LDA & AFT   & NFS   & TTG  & GRFG & GRFG$^\upsilon$ & GRFG$^\alpha$          \\  \midrule
Higgs Boson        & UCIrvine & C  & 50000   & 28       & 0.683 & 0.674 & 0.509 & 0.711 & 0.715 & 0.705 & 0.719 & 0.719 & \textbf{0.723} \\  
Amazon Employee    & Kaggle   & C   & 32769   & 9        & 0.744 & 0.740 & 0.920  & 0.943 & 0.935 & 0.806 & \textbf{0.946}  & 0.945 & 0.946\\  
PimaIndian         & UCIrvine & C   & 768     & 8        & 0.693 & 0.703 & 0.676  & 0.736 & 0.762 & 0.747 & 0.776  & 0.774 & \textbf{0.789}\\  
SpectF             & UCIrvine & C   & 267     & 44       & 0.790 & 0.748 & 0.774  & 0.775 & 0.876 & 0.788 & \textbf{0.878}  & 0.832 & 0.876 \\  
SVMGuide3  & LibSVM & C & 1243 & 21 & 0.703 & 0.747 & 0.683 & 0.829 & 0.831 & 0.766 & 0.850 & 0.852 & \textbf{0.858}\\  
German Credit      & UCIrvine & C   & 1001    & 24       & 0.695 & 0.661 & 0.627  & 0.751 & 0.765 & 0.731 & 0.772  & 0.773 & \textbf{0.774}\\  
Credit Default     & UCIrvine & C   & 30000   & 25       & 0.798 & 0.752 & 0.744  & 0.799 & 0.799 & \textbf{0.809} & 0.800  & 0.800 & 0.802 \\  
Messidor\_features & UCIrvine & C   & 1150    & 19       & 0.673 & 0.635 & 0.580  & 0.678 & 0.746 & 0.726 & 0.757  & 0.753 & \textbf{0.769} \\  
Wine Quality Red   & UCIrvine & C   & 999     & 12       & 0.599 & 0.611 & 0.600  & 0.658 & 0.666 & 0.647 & 0.686  & 0.688 & \textbf{0.697}\\  
Wine Quality White & UCIrvine & C   & 4900    & 12       & 0.552 & 0.587 & 0.571  & 0.673 & 0.679 & 0.638 & 0.685  & 0.685 & \textbf{0.693}\\  
SpamBase           & UCIrvine & C   & 4601    & 57       & 0.951 & 0.931 & 0.908  & 0.951 & 0.955 & \textbf{0.961}  & 0.958 & 0.952 & 0.958\\  
AP-omentum-ovary            & OpenML & C   & 275    & 10936        & 0.711 & 0.705 & 0.117  & 0.783 & 0.804 & 0.795 & 0.818  & 0.830 &\textbf{ 0.841}\\  
Lymphography       & UCIrvine & C   & 148     & 18       & 0.654 & 0.638 & 0.737 & 0.833 & 0.859 & 0.846 & \textbf{0.866}  & 0.866 & 0.863 \\  
Ionosphere         & UCIrvine & C   & 351     & 34       & 0.919 & 0.926 & 0.730  & 0.827 & 0.949 & 0.938 & 0.960  & 0.954 &\textbf{ 0.962}\\  \midrule
Bikeshare DC       & Kaggle   & R   & 10886   & 11       & 0.483 & 0.571 & 0.494  & 0.670 & 0.675 & 0.659 & 0.681  & 0.682 & \textbf{0.682}\\  
Housing Boston     & UCIrvine & R   & 506     & 13       & 0.605 & 0.617 & 0.174 & 0.641 & 0.665 & 0.658 & 0.684  & 0.668 & \textbf{0.692}\\  
Airfoil            & UCIrvine & R   & 1503    & 5        & 0.737 & 0.732 & 0.463  & 0.774 & 0.771 & 0.783 & 0.797  & 0.786 & \textbf{0.796}\\  
Openml\_618        & OpenML   & R   & 1000    & 50       & 0.415 & 0.427  & 0.372 & 0.665 & 0.640 & 0.587 & 0.672  & 0.704 & \textbf{0.803}\\  
Openml\_589        & OpenML   & R   & 1000    & 25       & 0.638 & 0.560 & 0.331  & 0.672 & 0.711 & 0.682 & 0.753  & 0.771 & \textbf{0.782}\\  
Openml\_616        & OpenML   & R   & 500     & 50       & 0.448 & 0.372 & 0.385  & 0.585 & 0.593 & 0.559 & 0.603  & 0.682 & \textbf{0.717}\\  
Openml\_607        & OpenML   & R   & 1000    & 50       & 0.579 & 0.406 & 0.376  & 0.658 & 0.675 & 0.639 & 0.680  & 0.702 & \textbf{0.756}\\  
Openml\_620        & OpenML   & R   & 1000    & 25       & 0.575 & 0.584 & 0.425 & 0.663 & 0.698 & 0.656 & 0.714  & 0.710 & \textbf{0.720}\\  
Openml\_637        & OpenML   & R   & 500     & 50       & 0.561 & 0.497 & 0.494  & 0.564 & 0.581 & 0.575 & 0.589  & 
0.632 & \textbf{0.644}\\  
Openml\_586        & OpenML   & R   & 1000    & 25       & 0.595 & 0.546 & 0.472  & 0.687 & 0.748 & 0.704 & 0.783  & 0.791 & \textbf{0.802}\\  \midrule
Mammography & UCIrvine   & D & 278 & 30 & 0.753 & 0.766 & 0.736  & 0.743 & 0.755 & 0.752 & 0.785 & 0.785 & \textbf{0.832} \\  
Yeast & OpenML   & D   & 11183    & 6       & 0.731 & 0.728 & 0.668  & 0.714 & 0.728 & 0.734 & 0.751 &\textbf{ 0.757} & 0.753 \\  
Wisconsin-Breast Cancer & UCIrvine   & D   & 3772    & 6       & 0.813 & 0.790 & 0.778  & 0.797 & 0.722 & 0.720 & 0.954   & 0.954 & \textbf{0.979}\\  
Thyroid & UCIrvine   & D   & 1364    & 8       & 0.852 & 0.862 & 0.589  & 0.873 & 0.847 & 0.832 & 0.979  & 0.984 & \textbf{0.998}\\  
SMTP  & UCIrvine   & D   & 95156    & 3       & 0.885 & 0.836 & 0.765  & 0.881 & 0.816 & 0.895 & 0.943  & 0.942 & \textbf{0.949}\\  
\bottomrule\\
\textbf{Average Ranking }& - & - & - & - & 7.05 & 7.39 & 8.58 & 5.65 & 4.63 & 5.37 & 2.48 & 2.50 & \textbf{1.37}\\
\\

\bottomrule
\end{tabular}}}
\vspace{-0.4cm}
\end{table*}

\subsubsection{Overall Comparison}
This experiment aims to answer:
\textit{Can our method effectively construct quality feature space and improve a downstream task?} 
The empirical results from our exhaustive experimental comparison provide insightful revelations concerning the efficacy of various feature transformation methods benchmarked over 29 diverse datasets. The nine methods evaluated in this study encompass a broad spectrum of techniques, ranging from random-based methods, expansion-reduction techniques, and matrix factorization approaches to reinforcement learning-based methods.
Table \ref{table_overall_perf} shows the comparison results in terms of F1 score, 1-RAE, or ROC/AUC.
We observed that GRFG and its variants (i.e., GRFG$^\upsilon$ and GRFG$^\alpha$)  rank first on most datasets and ranks second on ``Credit Default'' and ``SpamBase''.
The underlying driver is that the personalized feature-crossing strategy in GRFG considers feature-feature distinctions when generating new features.
We can also observe that GRFG$^\alpha$ beat the other two Q-learning approaches with an average ranking of 1.37. 
The underlying driver is that the Actor-Critic optimization perspective can learn the policy directly and separately learn the value function and policy evaluation, thus being more effective in the downstream tasks.
The average rankings of the remaining six baseline methods are substantially higher, indicating that these methods are generally less effective on the tested datasets. LDA, a matrix factorization-based method, ranks the lowest with an average rank of 8.58, suggesting that it may struggle with the challenges posed by the diverse datasets in our study. RDG and ERG, representing random-based and expansion-reduction methods, achieve rankings of 7.05 and 7.39, respectively. The relatively lower performance of RDG underscores the limitations of random-based feature generation, while the simple nature of the expansion-reduction technique might hinder ERG's performance.
AFT, an enhanced version of ERG, showcases an improvement over its simpler counterpart with an average ranking of 5.65. This improvement reinforces the value of additional enhancements in the basic expansion-reduction technique. The other two reinforcement learning-based methods, NFS and TTG, deliver average rankings of 4.63 and 5.37, respectively. These results signify that reinforcement learning methodologies have a compelling potential for feature transformation tasks, even though their performance lags behind our GRFG variants. 
Besides, the observation that GRFG outperforms random-based  (RDG) and expansion-reduction-based (ERG, AFT) methods 
shows that the agents can share states and rewards in a cascading fashion and, thus, learn an effective policy to select optimal crossing features and operations.
Also, we observed that GRFG could considerably increase downstream task performance on datasets with fewer features (e.g., Higgs Boson and SMTP) or datasets with more extensive features  (e.g., AP-omentum-ovary). These findings demonstrated the generalizability of GRFG.
Moreover, because our method is a self-learning end-to-end framework, users can treat it as a tool and easily apply it to different datasets regardless of implementation details.
Thus, this experiment validates that our method is more practical and automated in real application scenarios.

\begin{figure*}[!t]
\vspace{-0.1cm}
\centering
\subfigure[SVMGuide3]{
\includegraphics[width=3.4cm]{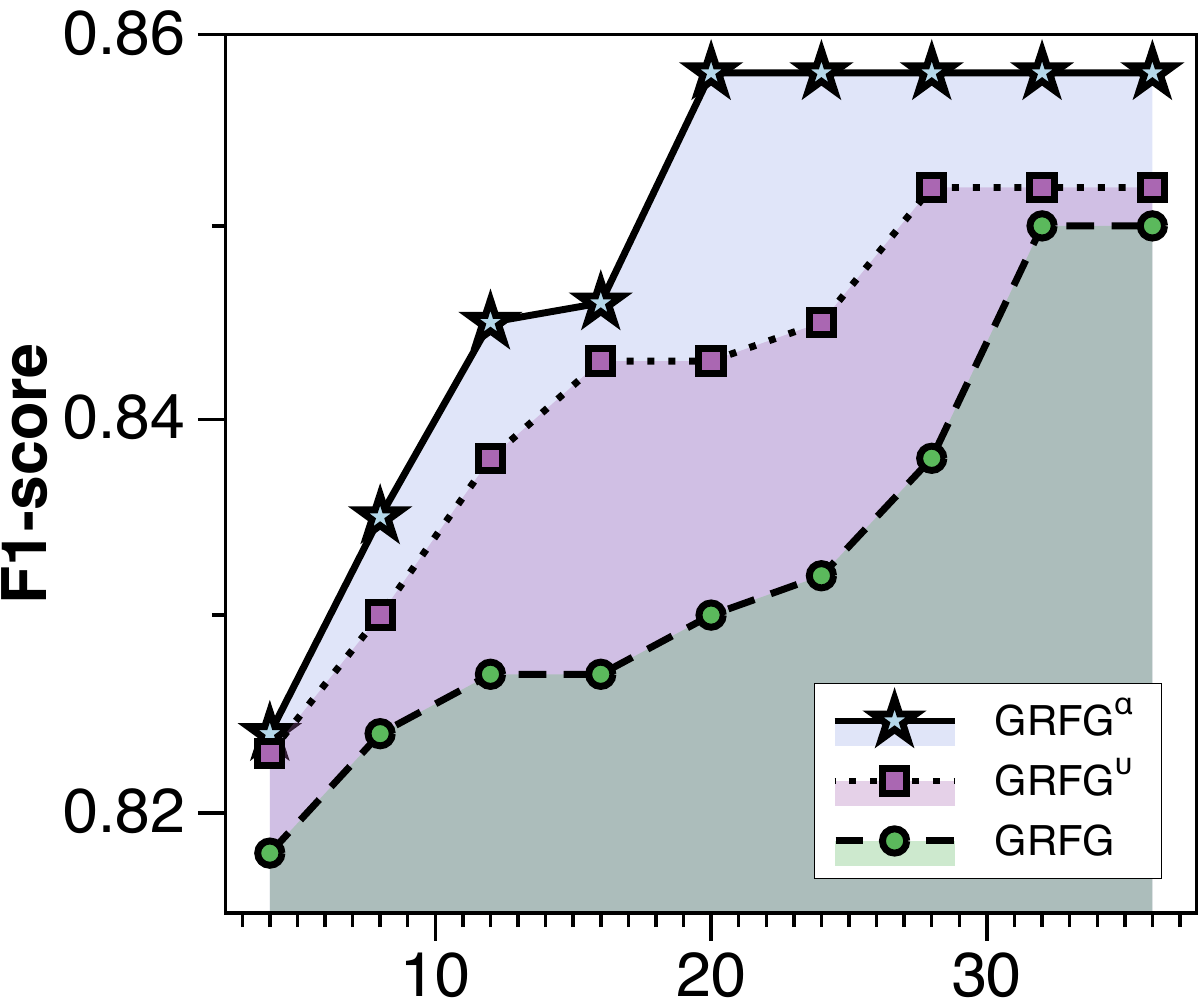}
}
\hspace{-3mm}
\subfigure[OpenML\_586]{ 
\includegraphics[width=3.4cm]{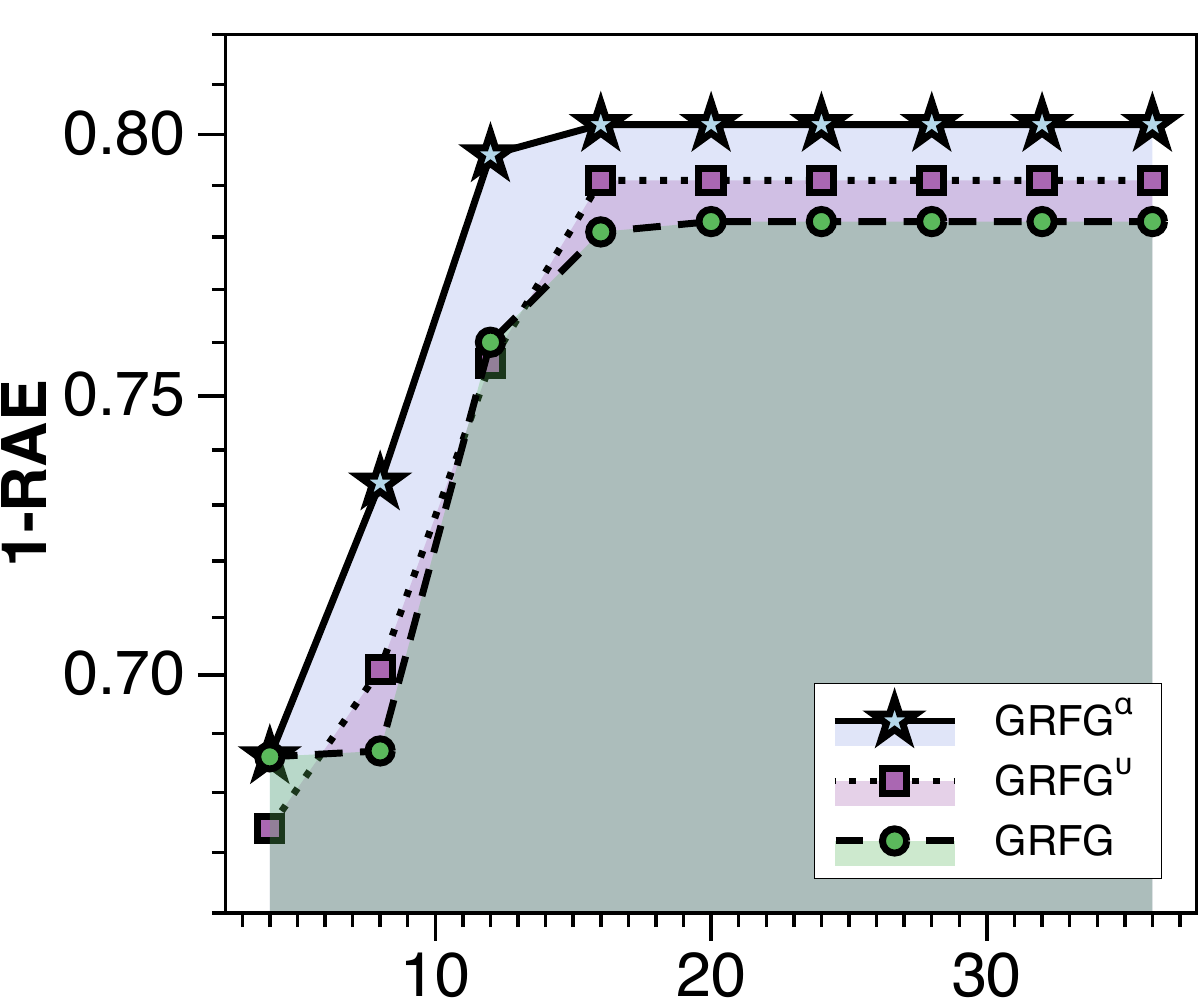}
}
\hspace{-3mm}
% % \vspace{-4mm}
\subfigure[OpenML\_589]{
\includegraphics[width=3.4cm]{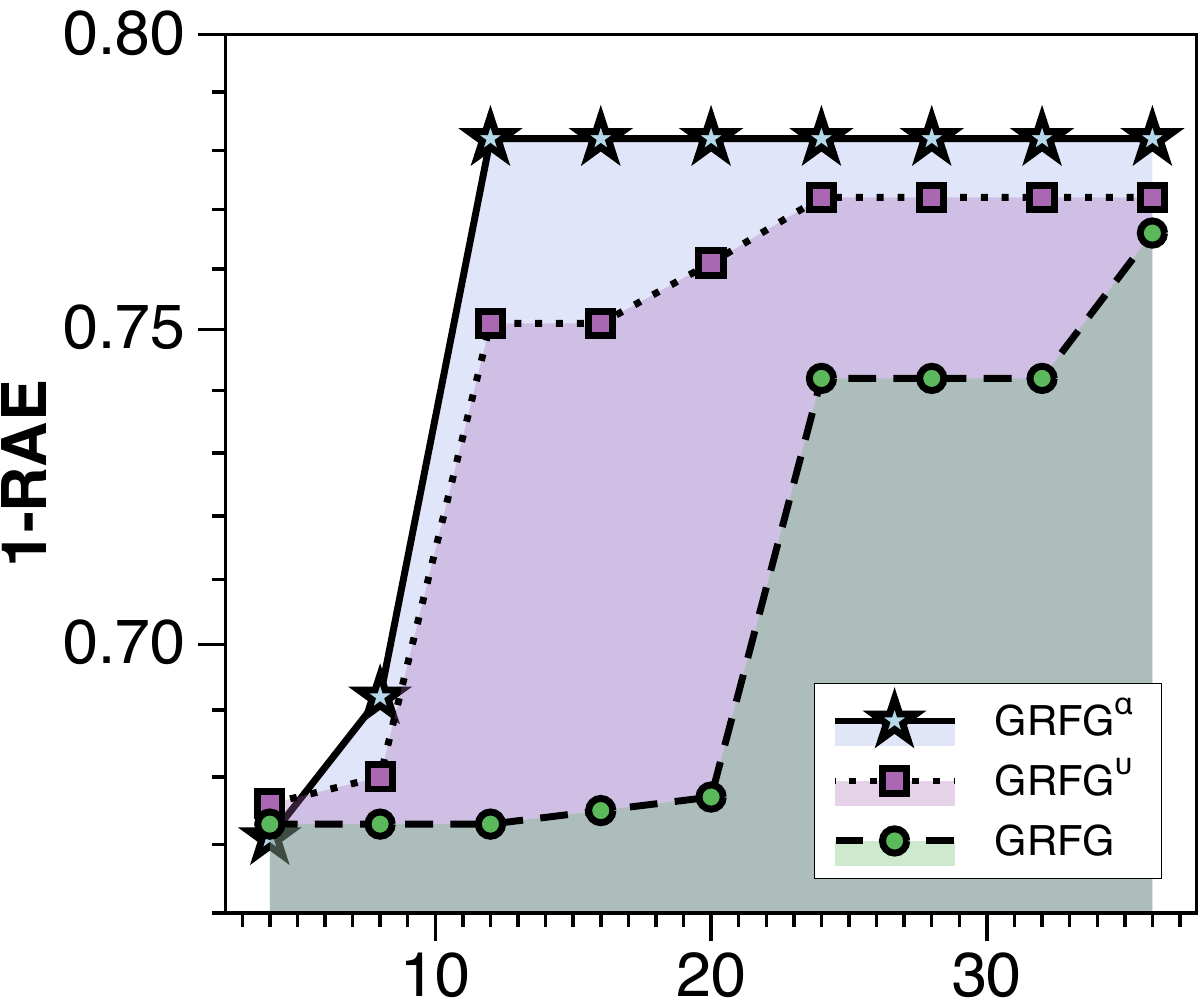}
}
\hspace{-3mm}
\subfigure[Thyroid]{ 
\includegraphics[width=3.4cm]{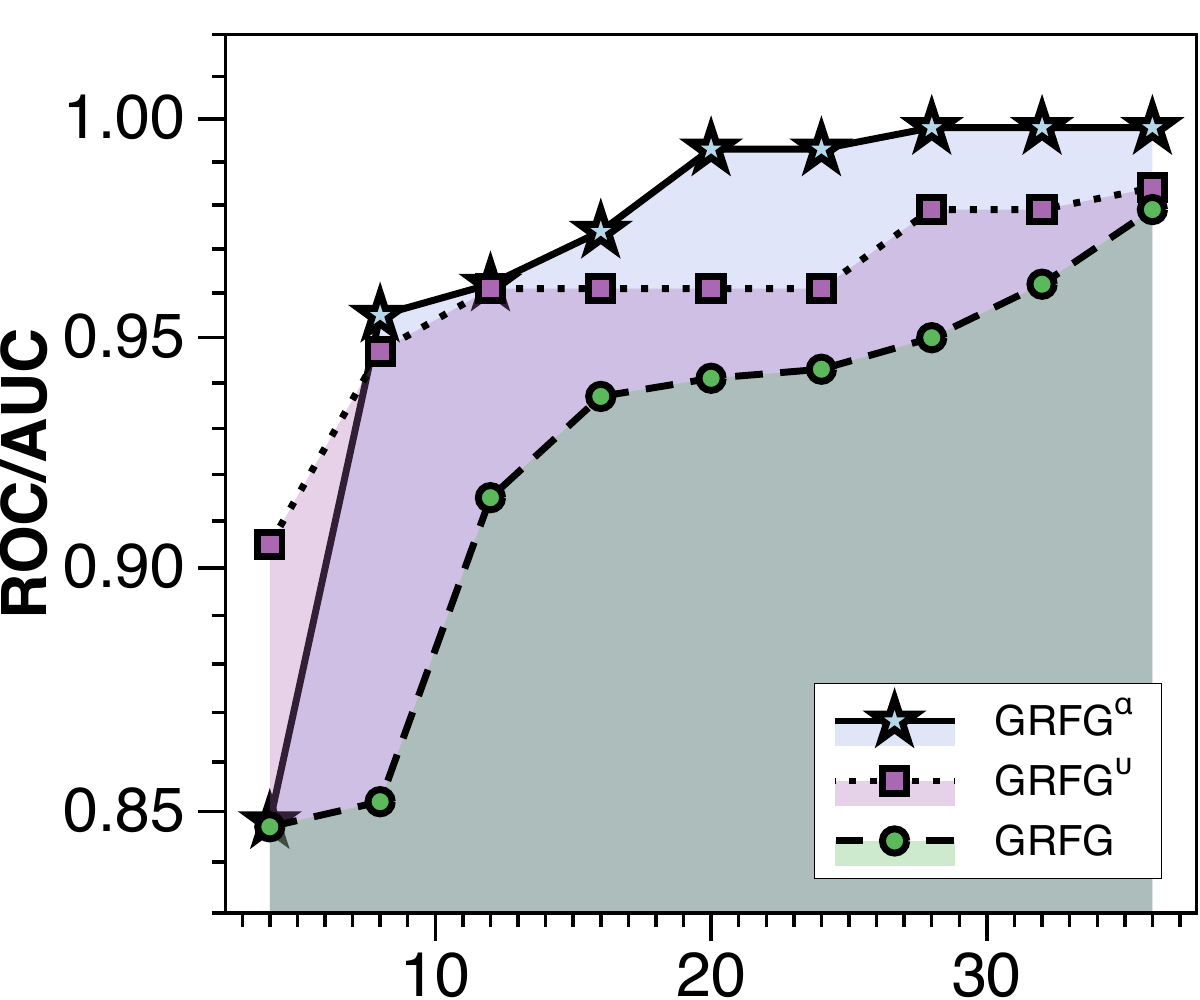}
}
\vspace{-0.35cm}
\caption{Comparison of different optimization perspectives in different tasks.}
\label{rl_study}
\vspace{-0.4cm}
\end{figure*}

\subsubsection{Study of the impact of improving Q-learning}
This experiment aims to answer: \textit{How does the improving Q-learning variant of GRFG affect the feature space reconstruction?}
In Section~\ref{ot}, we decoupled the action selection network with the Q-value estimation network and the state value function with the action advantage function, thus improving the Q-value over-estimating. From those perspectives above, we modified the original framework to GRFG$^\upsilon$, an improving Q-learning method. From Table~\ref{table_overall_perf}, we can observe that GRFG$^\upsilon$ ranks similarly to the original model, delivering average rankings of 2.50 and 2.48, respectively. However, from Figure~\ref{rl_study}, we can observe that GRFG$^\upsilon$ will converge much more effectively than the original model. For example, in OpenML\_589,  GRFG$^\upsilon$ can converge in epoch 24, but GRFG requires around 36 epochs. The original GRFG was observed to lag behind its counterparts. Despite DQN's renown in addressing high-dimensional and continuous state spaces, it seemed less capable in this specific application, potentially due to its relative propensity for overestimation bias. This limitation is partially mitigated in the architecture of GRFG$^\upsilon$, which aims to separate the estimator into two streams for estimating the state value and the advantages for each action, hence reducing over-optimistic value estimates.
Indeed, the GRFG$^\upsilon$ framework achieves more efficient convergence than the baseline GRFG model, even though the final performances of the two models were found to be equivalent. This outcome reaffirms the benefits of the architecture, which separates the estimation of state value and advantage for each action. Consequently, this results in more stable learning and faster convergence.

\subsubsection{Study of the impact of the Actor-Critic optimization perspective}
This experiment aims to answer: \textit{How does the Actor-Critic optimization perspective affect the feature space reconstruction?} 
To answer this question, we develop a novel variant of GRFG, denoted by GRFG$^\alpha$, which utilizes an Actor-Critic optimization perspective proposed in this journal version. 
Among all baseline methods in Table~\ref{table_overall_perf}, our novel GRFG variants incorporating the Actor-Critic optimization strategy exhibit a strong competitive edge over the other baselines. As per the average rankings over the datasets, GRFG$^\alpha$ attains the highest average rank of 1.37, outperforming the original GRFG (Q-learning) and GRFG$^\upsilon$ (improving Q-learning) with average rankings of 2.48 and 2.50, respectively. 
Besides, from Figure~\ref{rl_study}, the GRFG$^\alpha$ framework was found to surpass the two others in terms of both convergence efficiency and achieved performance. 
% This suggests that the Actor-Critic approach, which simultaneously learns a policy function and a value function, enhances the general capabilities of the GRFG framework.
The superior performance of GRFG$^\alpha$ could be attributed to the Actor-Critic method's ability to balance the trade-off between exploration and exploitation more effectively than the other two variants, accelerating the convergence process and improving the overall result.
This evidences the superior effectiveness of the Actor-Critic framework within the GRFG context, as it enables a more effective balance between exploration and exploitation, resulting in superior performance.
These observations underline the importance of carefully selecting and tailoring reinforcement learning methods to specific tasks. 
While DQN has been famous for its ability to handle complex tasks, it may not always be the optimal solution. 
In this case, an Actor-Critic approach (GRFG$^\alpha$) resulted in more efficient convergence and superior performance. 
Yet, it is also important to note that the improving Q-learning method (GRFG$^\upsilon$) offered some advantages in terms of convergence over the standard DQN (GRFG), emphasizing the potential benefits of such refined Q-learning variants.

\begin{figure*}[!t]
\vspace{-0.1cm}
\centering
\subfigure[German Credit]{
\includegraphics[width=3.4cm]{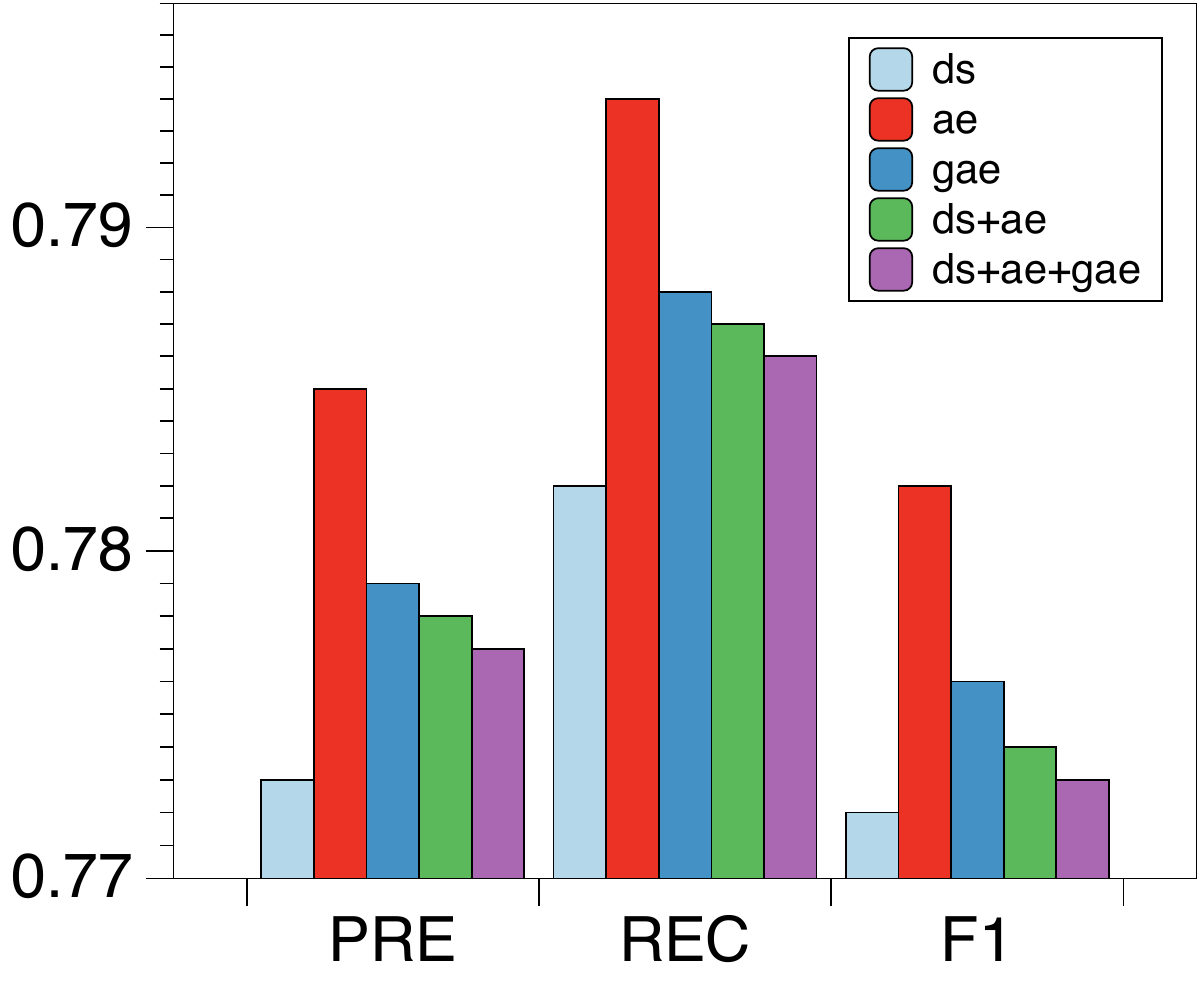}
}
\hspace{-3mm}
\subfigure[OpenML\_586]{ 
\includegraphics[width=3.4cm]{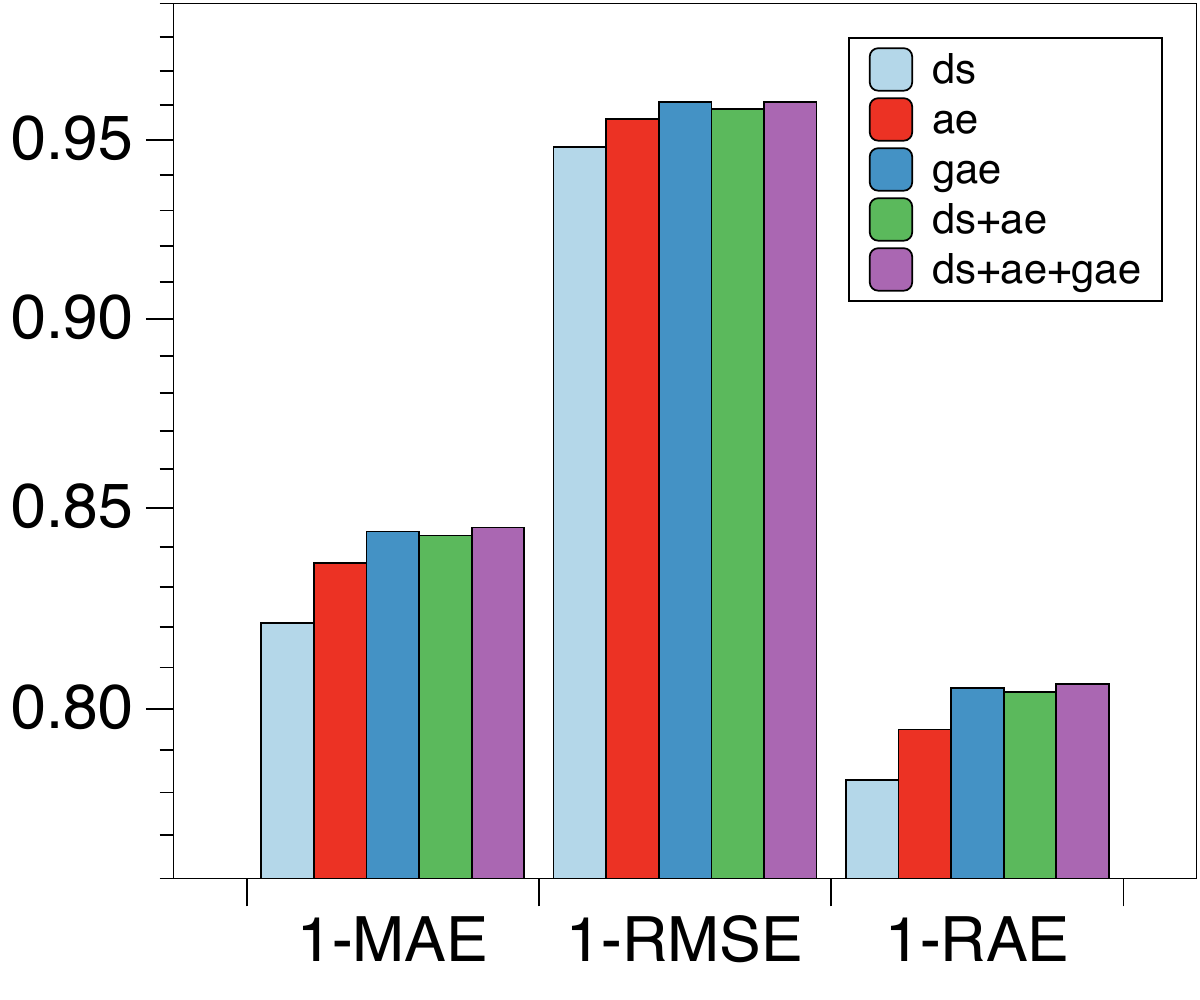}
}
\hspace{-3mm}
% % \vspace{-4mm}
\subfigure[OpenML\_589]{
\includegraphics[width=3.4cm]{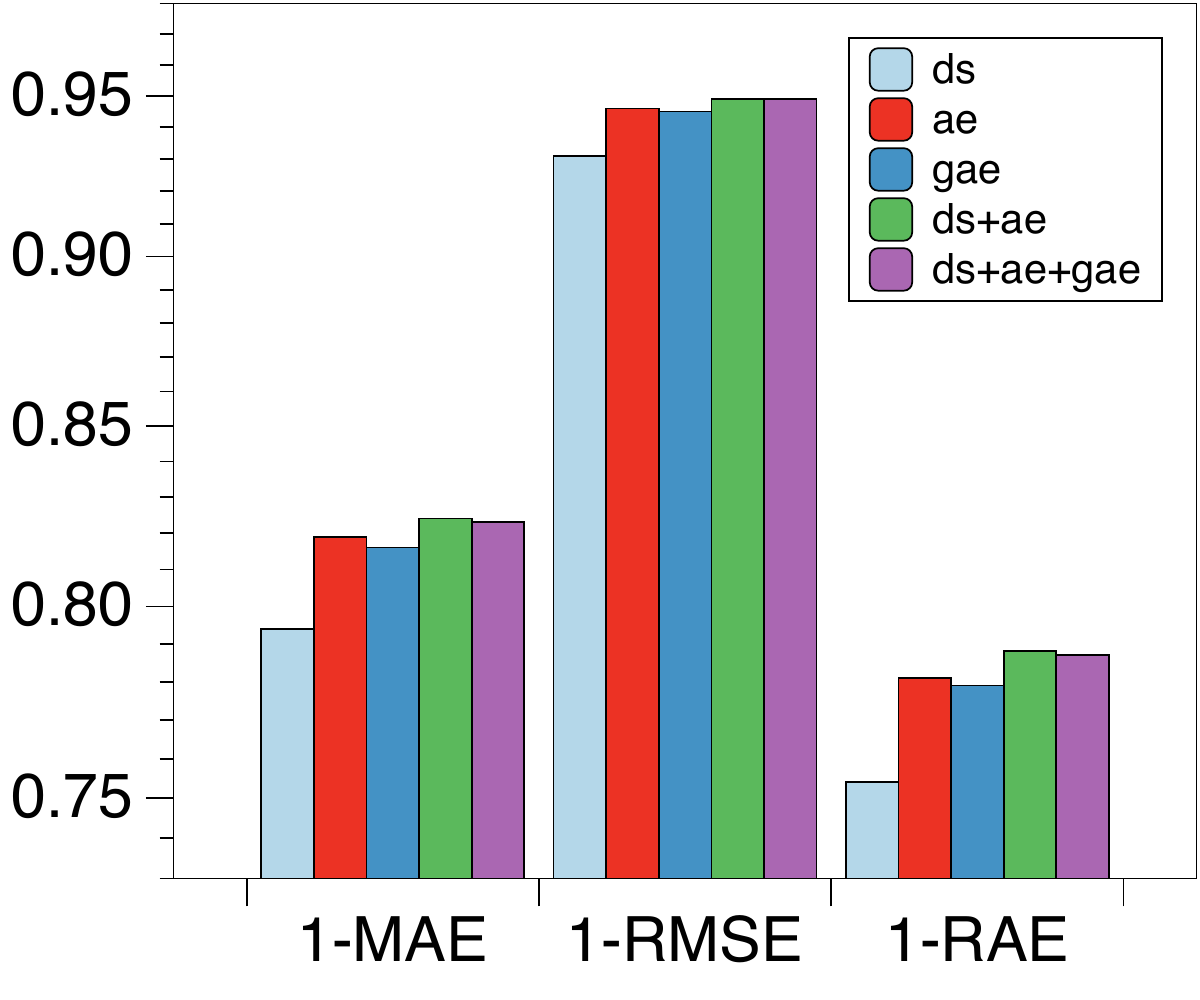}
}
\hspace{-3mm}
\subfigure[Thyroid]{ 
\includegraphics[width=3.4cm]{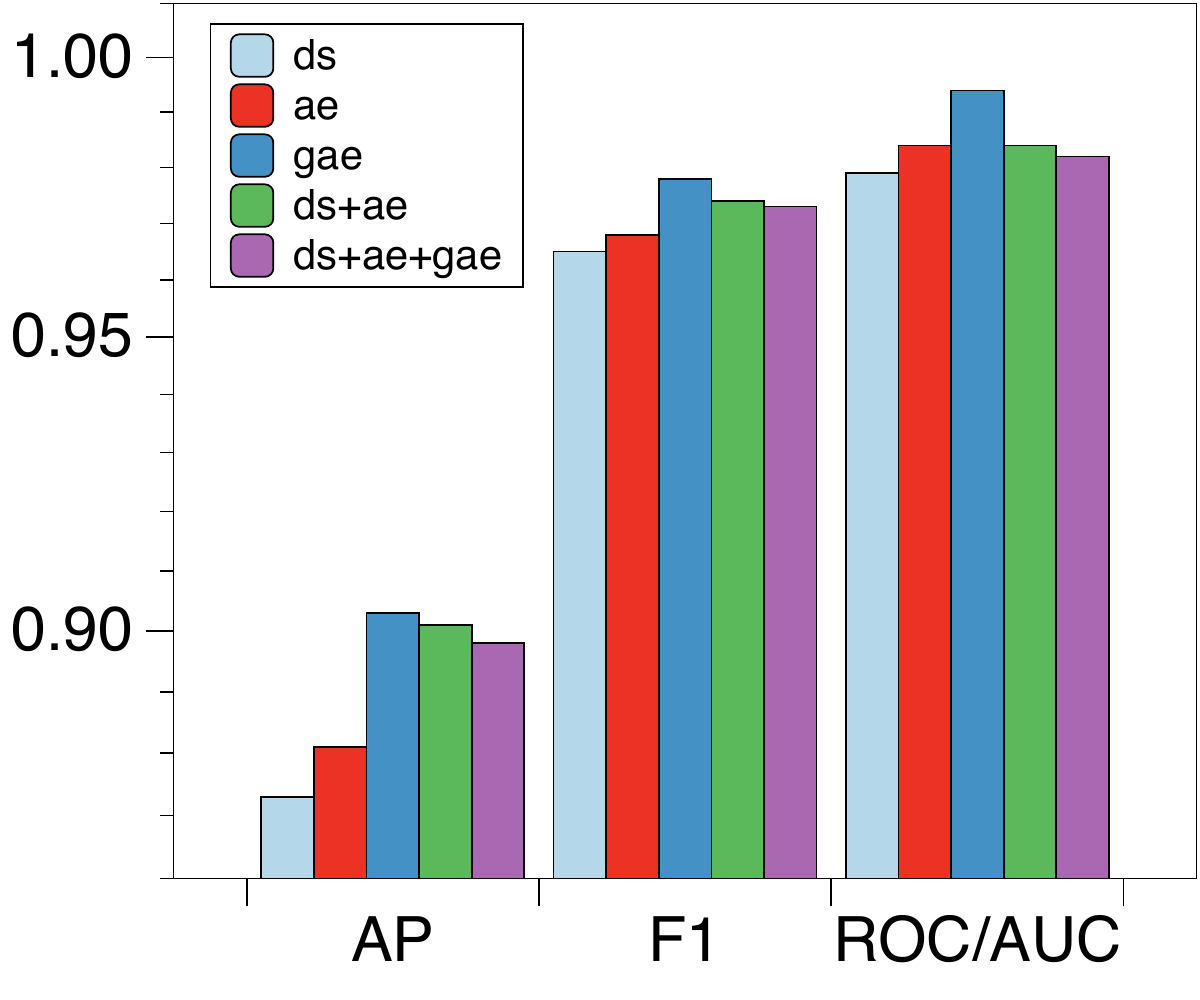}
}
\vspace{-0.35cm}
\caption{Comparison of different state representation methods in different tasks.}
\label{state_study}
\vspace{-0.4cm}
\end{figure*}

% For that, we use various reinforcement methods as the backbone of GRFG for testing the model generalization and efficiency in converage. We reported the experimental results in the Figure~\ref{rl_study}. The results show that although the randomness in reinforcement learning could affect the model performance in each epoch, the Dueling DQN and Dueling Double DQN converge much more efficiently in most tasks. For example, in SVMGuide3 tasks, the Dueling DQN can converge in epoch 12, but the DQN and DDQN require around 50 epochs to achieve the best performance. In the OpenML\_589 task, the Dueling Double DQN can achieve its best performance with only 6 epochs. However, the DQN needs to be trained in more than 100 epochs to achieve equivalent performance. We interpreted this as the Dueling DQN and Dueling Double DQN are more robust to learning the Q value, thus facilitating the model's convergence. 

\subsubsection{Study of different state representation methods}
 
This experiment aims to answer: \textit{
What effects does state representation have on improving feature space?}
In the Section~\ref{rep_fun},
we implemented three state representation methods: descriptive statistics based (\textbf{ds}), antoencoder based (\textbf{ae}), and graph autoencoder based (\textbf{gae}).
Additionally, we developed two mixed state representation strategies: \textbf{ds}+\textbf{ae} and \textbf{ds}+\textbf{ae}+\textbf{gae}.
Figure~\ref{state_study} shows the comparison results of classification tasks in terms of Precision (PRE), Recall (REC), and F1-score (F1); regression tasks in terms of 1 - Mean Average Error (1-MAE), 1 - Root Mean Square Error (1-RMSE), and 1 - relative absolute error (1-RAE);  outlier detection tasks in terms of Average Precision (AP), F1-score (F1), Area Under the Receiver Operating Characteristic Cur (ROC/AUC).
We found that advanced state representation methods  (\textbf{ae} and \textbf{gae}) outperform the method (\textbf{ds}) provided in our preliminary work in most cases.
For instance, in classification tasks (e.g., German Credit and SVMGuide3), the GRFG with \textbf{ae} outperform the preliminary version (i.e., the version with \textbf{ds}). In outlier detection tasks, the \textbf{gae} state representation method perform significantly better than the \textbf{ds} method. In regression tasks, two advanced methods achieve equivalent performance but considerably better than \textbf{ds}. 
A possible reason is that \textbf{ae} and \textbf{gae} capture more intrinsic feature properties through considering individual feature information and feature-feature correlations.
Another interesting observation is that 
mixed state representation strategies (\textbf{ds}+\textbf{ae} and \textbf{ds}+\textbf{ae}+\textbf{gae}) do not achieve the best performance on all tasks.
The underlying driver is that although concatenated state representations can include more feature information, redundant noises may lead to sub-optimal policy learning.

\subsubsection{Study of the impact of each technical component}
\label{study_lp}

\begin{figure*}[!t]
\vspace{-0.1cm}
\centering
\subfigure[PimaIndian]{
\includegraphics[width=3.4cm]{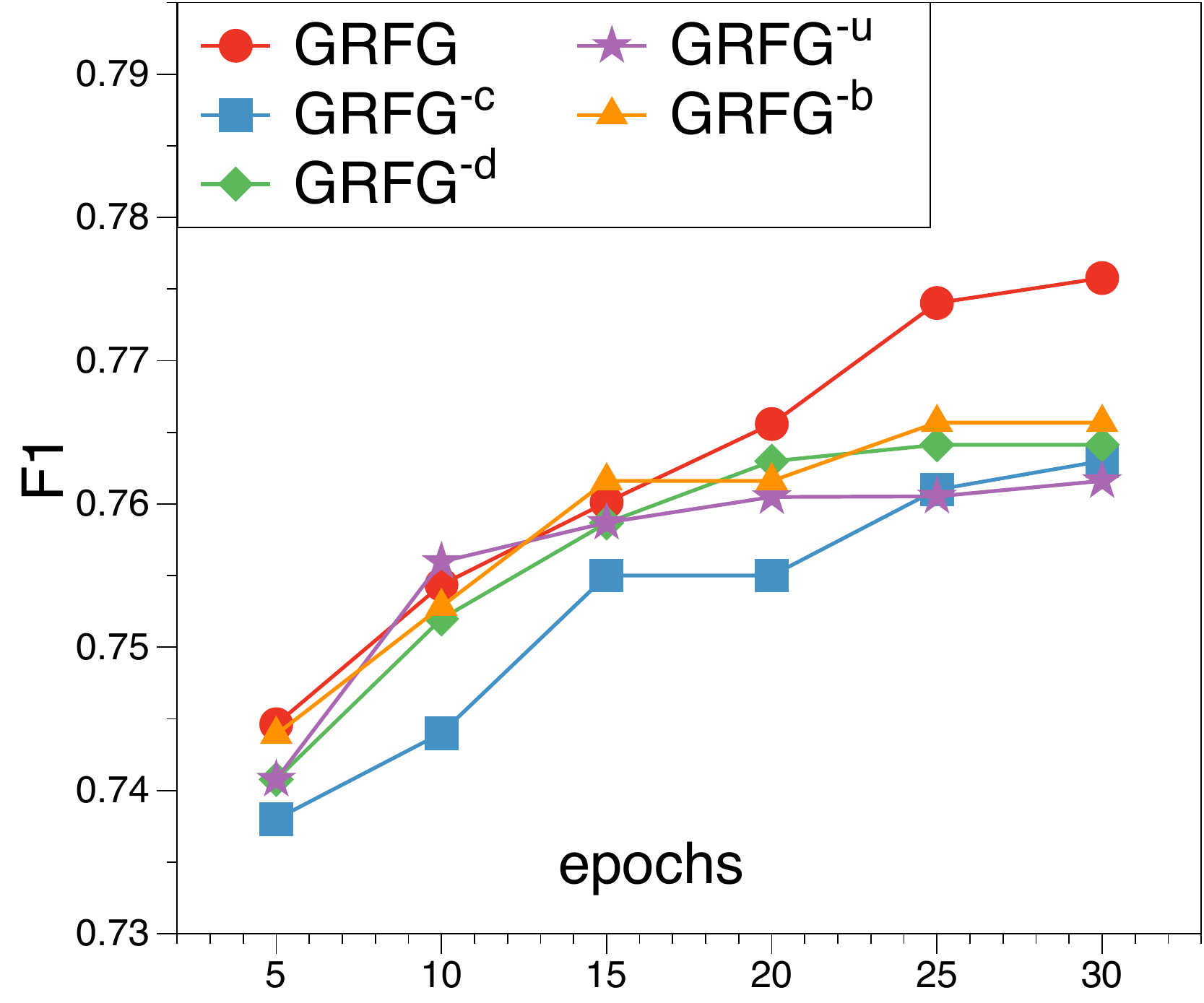}
}
\hspace{-3mm}
\subfigure[German Credit]{ 
\includegraphics[width=3.4cm]{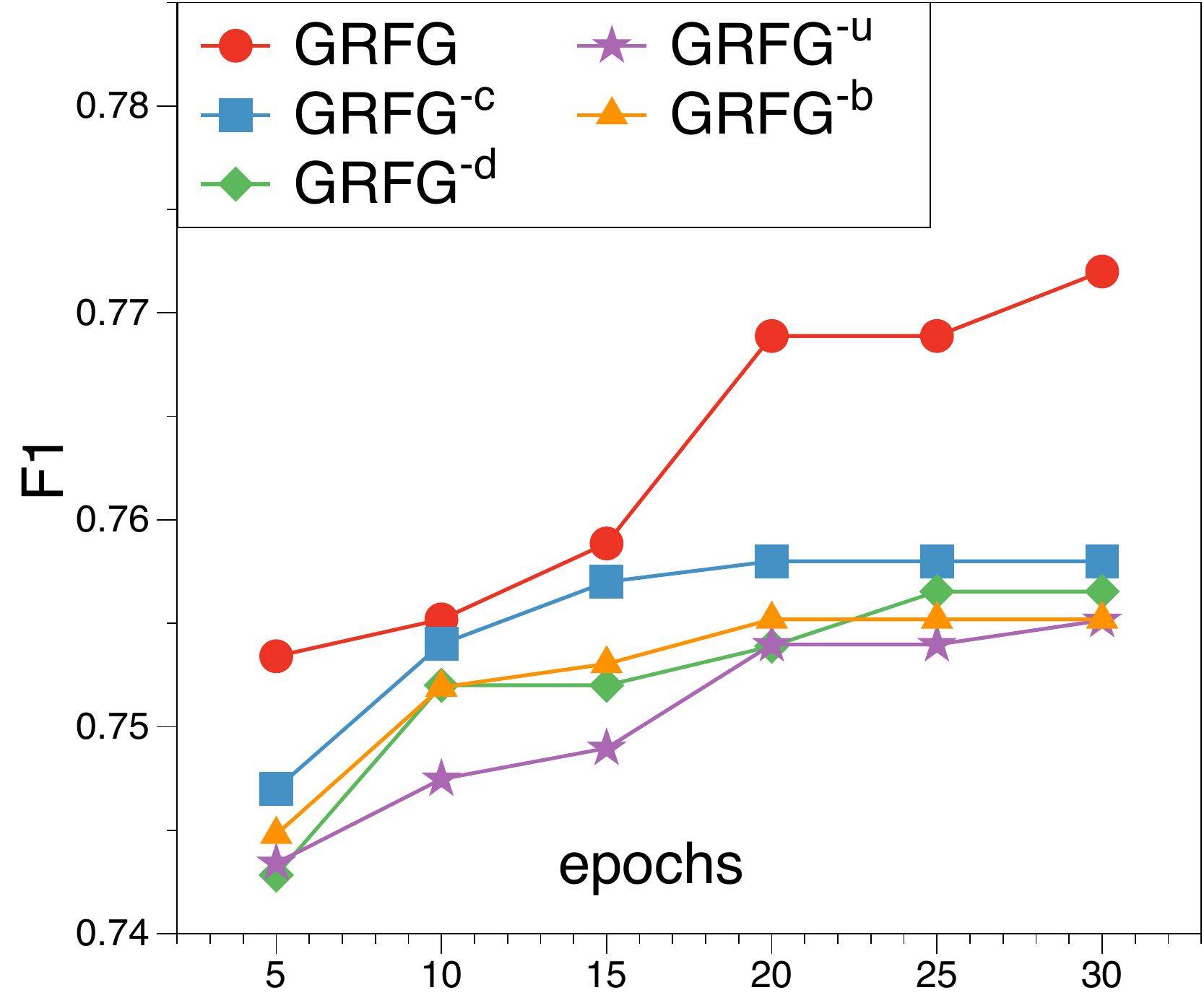}
}
\hspace{-3mm}
% % \vspace{-4mm}
\subfigure[Housing Boston]{
\includegraphics[width=3.4cm]{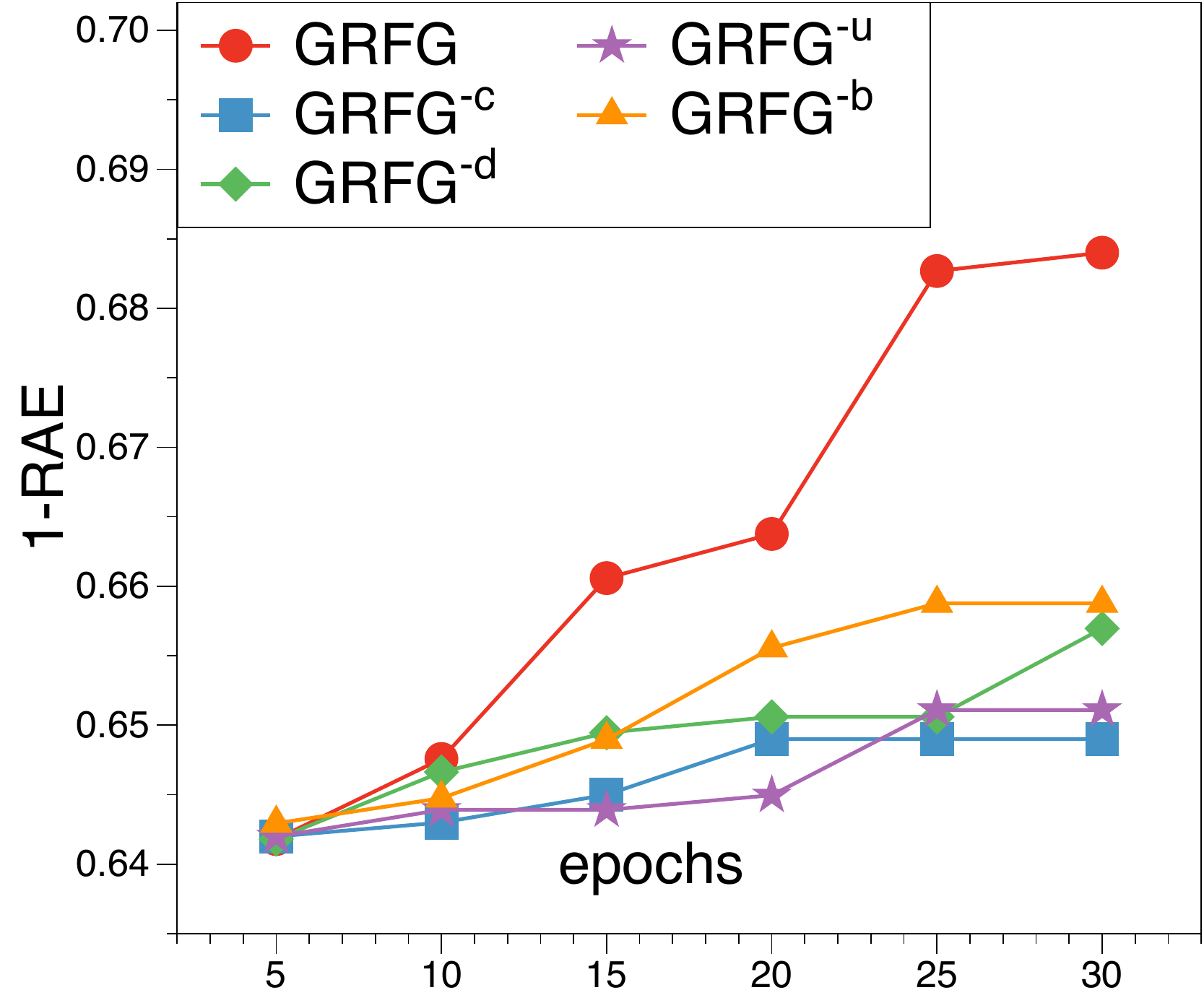}
}
\hspace{-3mm}
\subfigure[Openml\_589]{ 
\includegraphics[width=3.4cm]{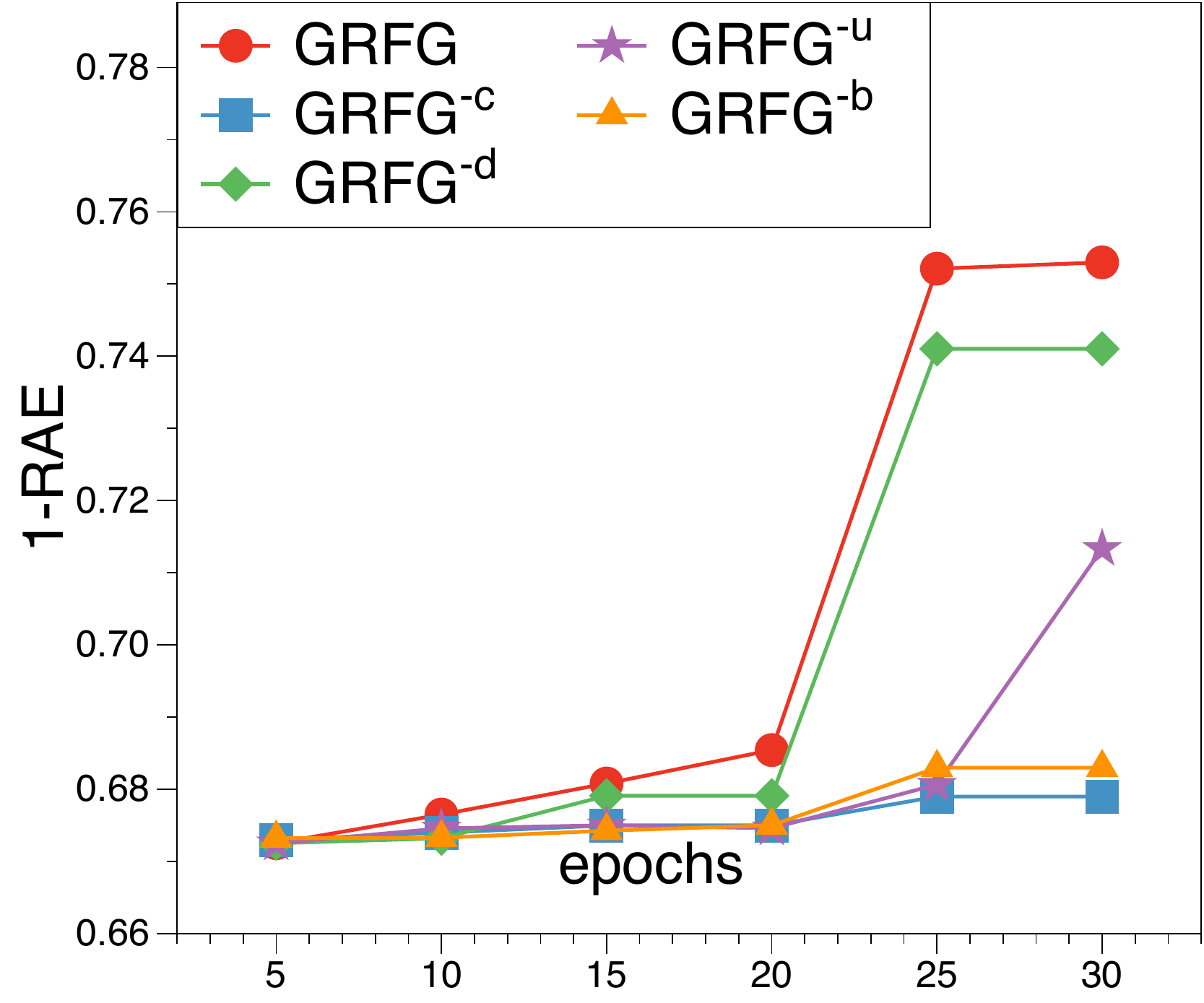}
}
\vspace{-0.35cm}
\caption{Comparison of different GRFG variants in terms of F1 or 1-RAE.}
\label{ab_study}
\vspace{-0.4cm}
\end{figure*}
This experiment aims to answer:
\textit{How does each component in GRFG impact its performance?}
We developed four variants of GRFG (Section \ref{baseline}). 
Figure ~\ref{ab_study} shows the comparison results in terms of F1 score or 1-RAE on two classification datasests (\textit{i.e.} PimaIndian, German Credit) and two regression datasets (\textit{i.e.} Housing Boston, Openml\_589). 
First, we developed GRFG$^{-c}$ by removing the feature clustering step of GRFG. 
But, GRFG$^{-c}$ performs poorer than GRFG on all datasets.
This observation shows that the idea of group-level generation can augment reward feedback to help cascading agents learn better policies. 
%This generation way augments the reward signals of cascading agents, enabling them to learn a more effective policy for feature group and operation selection.
Second, we developed  GRFG$^{-d}$ by using euclidean distance as feature distance metric in the M-clustering of GRFG.
The superiority of GRFG over GRFG$^{-d}$ suggests that our distance describes group-level information  relevance and redundancy ratio in order to maximize information distinctness across feature groups and minimize it within a feature group. Such a distance can help GRFG generate more useful dimensions.
%our proposed feature distance metric takes the information distinctness into account to produce more robust feature groups for feature generation.
Third, we developed GRFG$^{-u}$ and GRFG$^{-b}$ by using random in the two feature generation scenarios (Section \ref{group_generation}) of  GRFG.
We observed that GRFG$^{-u}$ and GRFG$^{-b}$ perform poorer than GRFG. This validates that crossing two distinct features and relevance prioritized generation can generate more meaningful and informative features.

\subsubsection{Study of the impact of M-Clustering}
\vspace{-0.1cm}
\begin{figure*}[!t]
\centering
\subfigure[PimaIndian]{
\includegraphics[width=3.4cm]{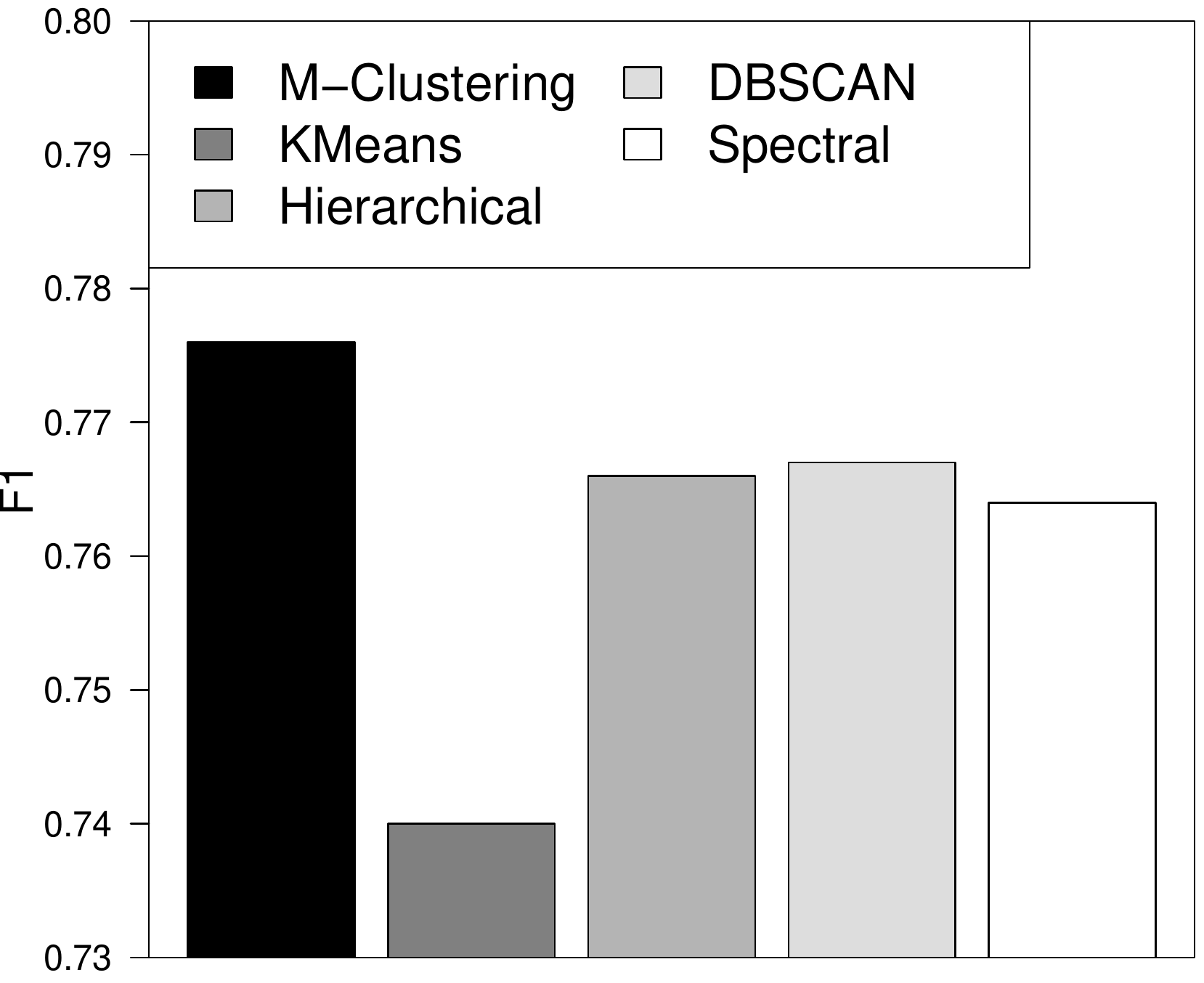}
}
\hspace{-3mm}
\subfigure[German Credit]{ 
\includegraphics[width=3.4cm]{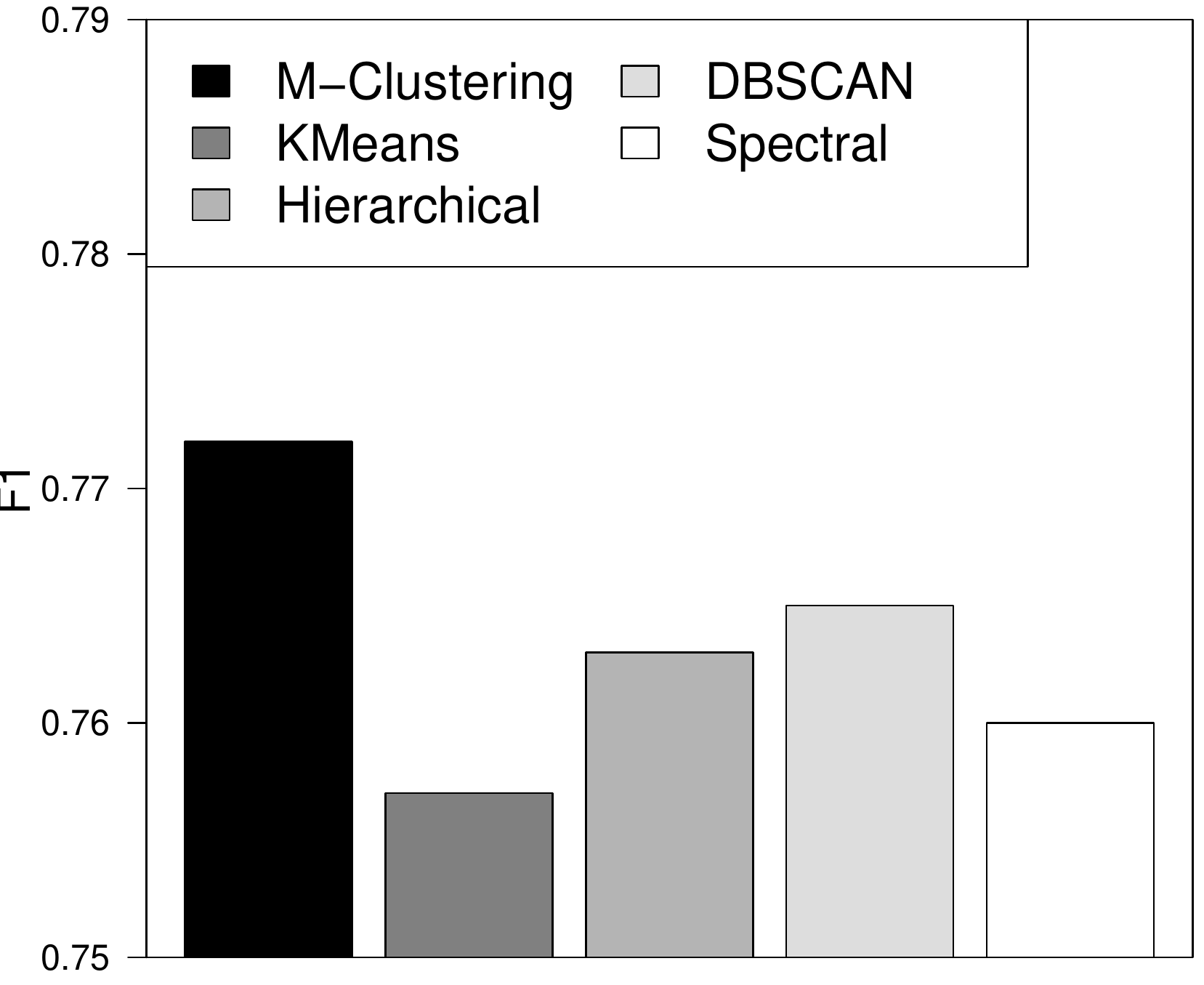}
}
\hspace{-3mm}
% % \vspace{-4mm}
\subfigure[Housing Boston]{
\includegraphics[width=3.4cm]{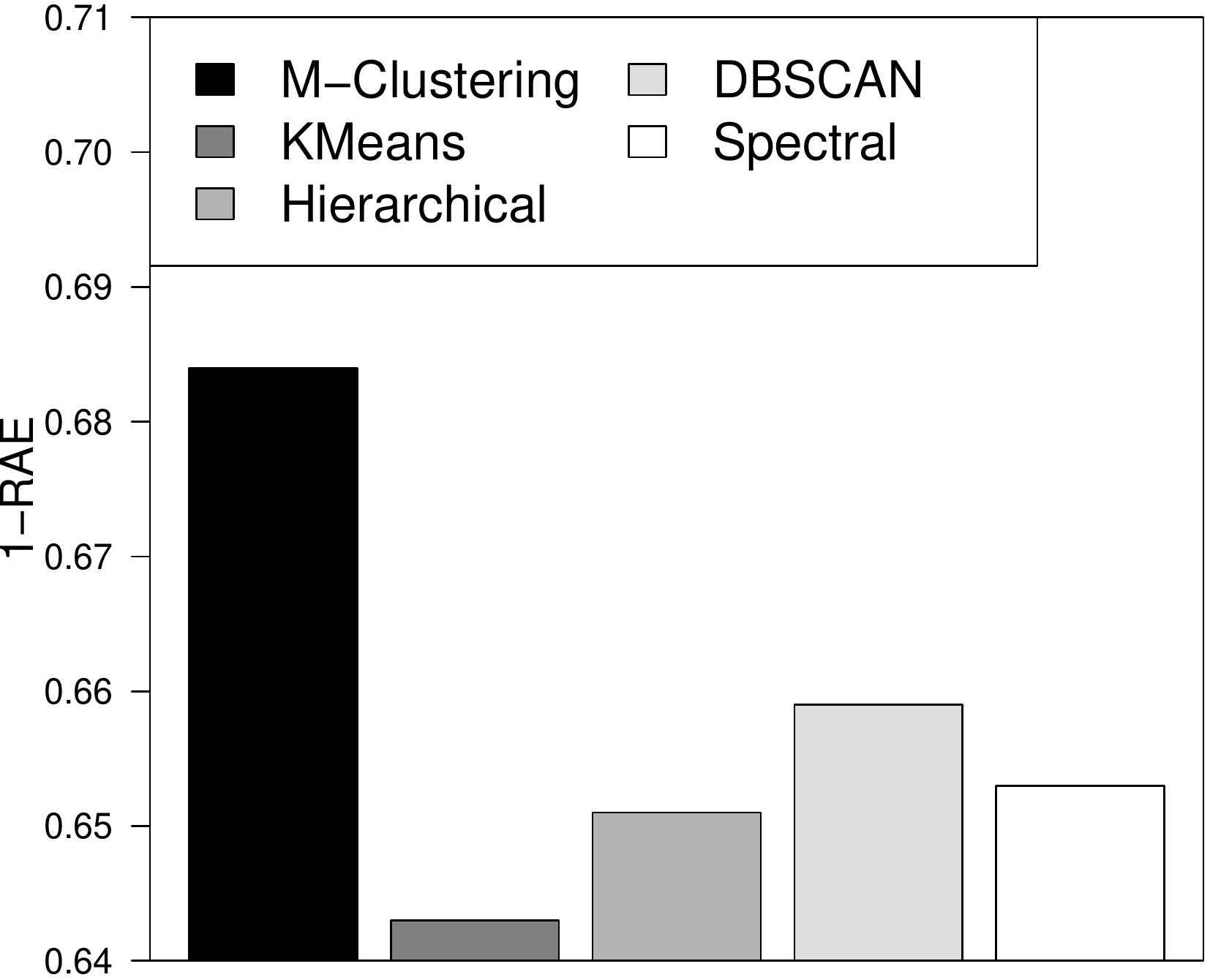}
}
\hspace{-3mm}
\subfigure[Openml\_589]{ 
\includegraphics[width=3.4cm]{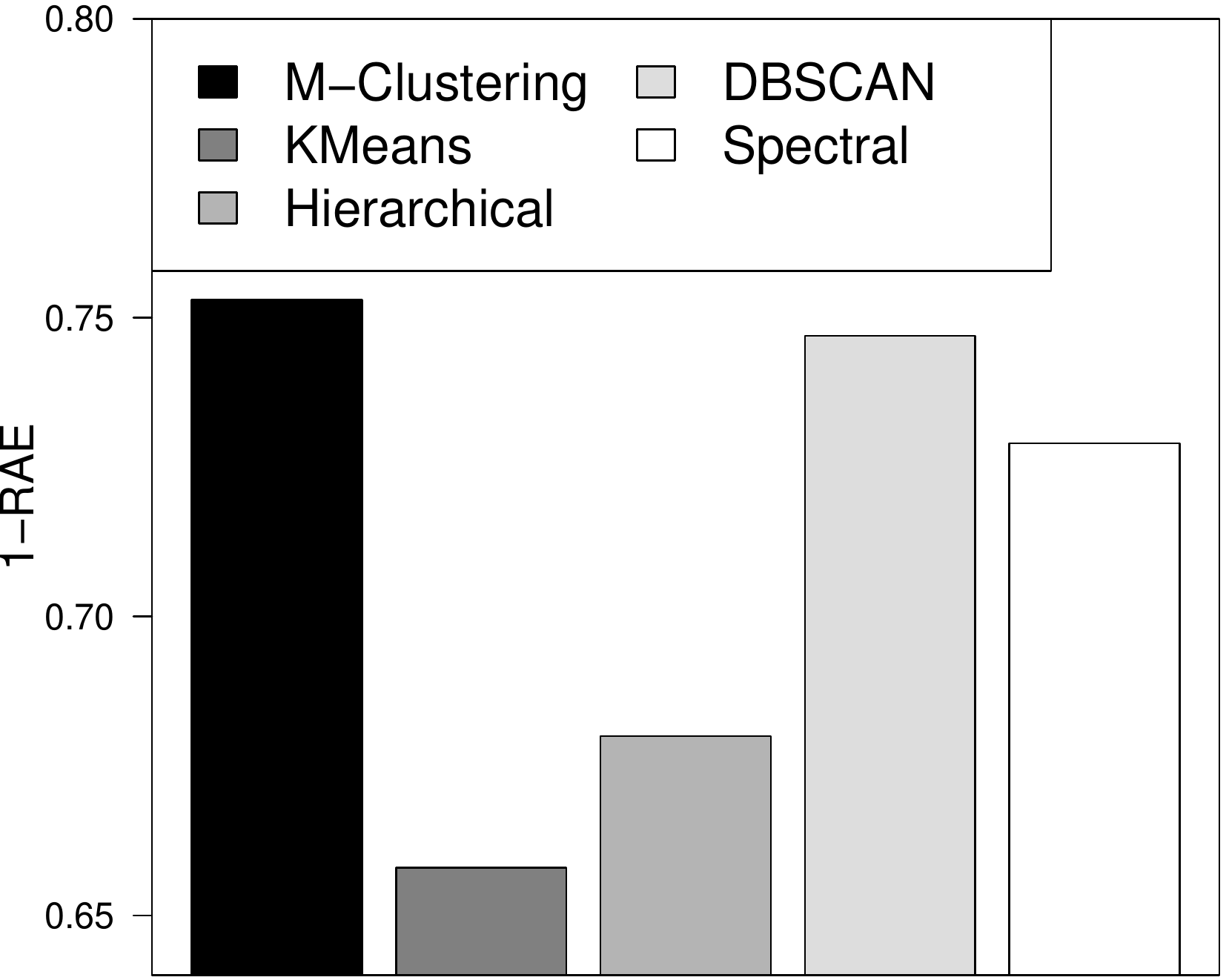}
}
\vspace{-0.35cm}
\caption{Comparison of different clustering algorithms in terms of F1 or 1-RAE.}
\label{differ_cluster}
\vspace{-0.35cm}
\end{figure*}
This experiment aims to answer: \textit{Is M-Clustering more effective in improving feature generation than classical clustering algorithms?}
We replaced the feature clustering algorithm in GRFG with KMeans, Hierarchical Clustering, DBSCAN, and Spectral Clustering respectively.
We reported the comparison results in terms of F1 score or 1-RAE on the datasets used in Section \ref{study_lp}.
Figure \ref{differ_cluster} shows M-Clustering beats classical clustering algorithms on all datasets.
The underlying driver is that when feature sets change during generation, M-Clustering is more effective in minimizing information overlap of intra-group features and maximizing information distinctness of  inter-group features. So, crossing  the feature groups with distinct information is easier to generate  meaningful dimensions.

\subsubsection{Study of the robustness of GRFG under different machine learning (ML) models.}
\begin{figure*}[!t]
% \vspace{-0.1cm}
\centering
\subfigure[PimaIndian]{
\includegraphics[width=3.4cm]{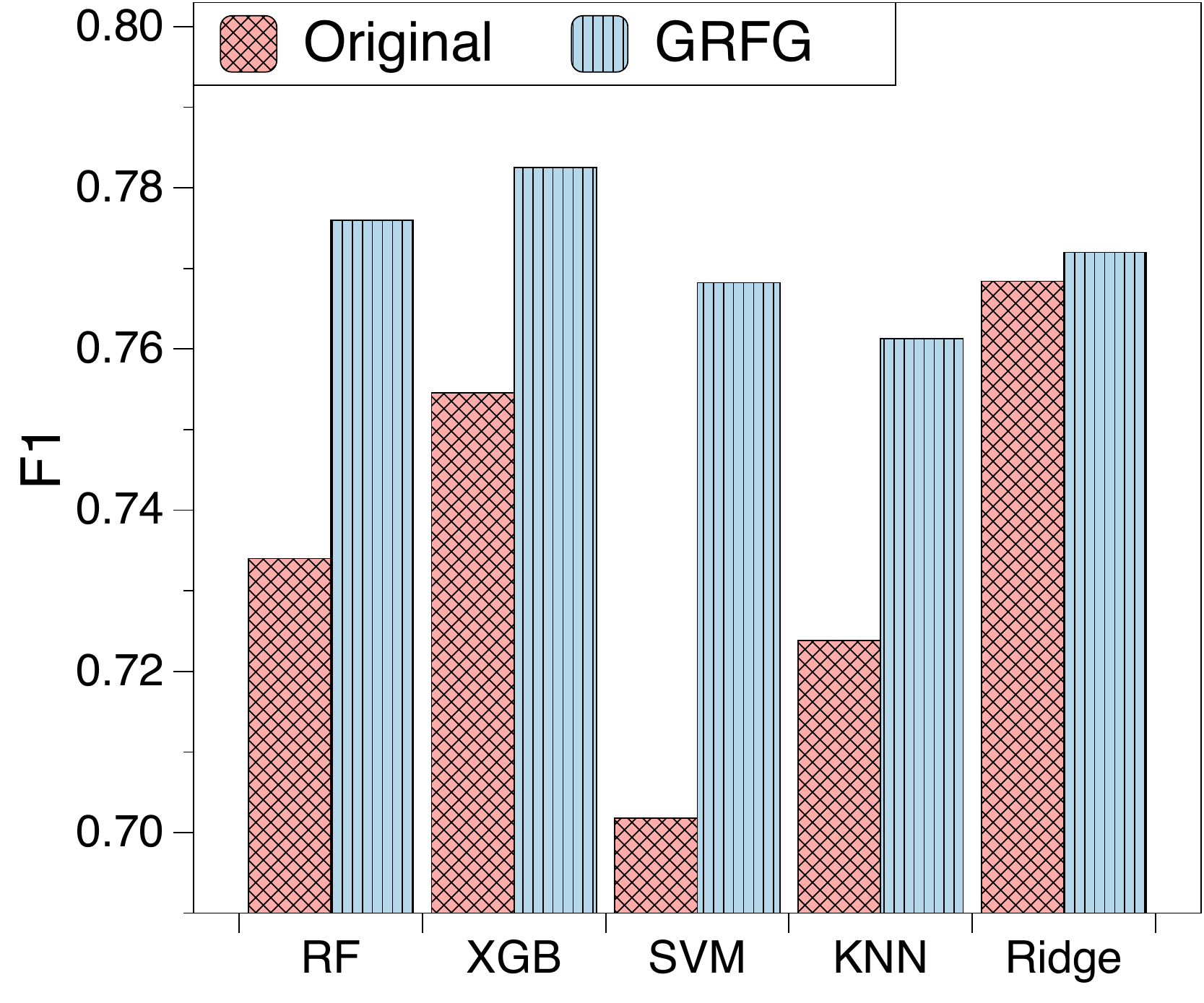}
}
\hspace{-3mm}
\subfigure[German Credit]{ 
\includegraphics[width=3.4cm]{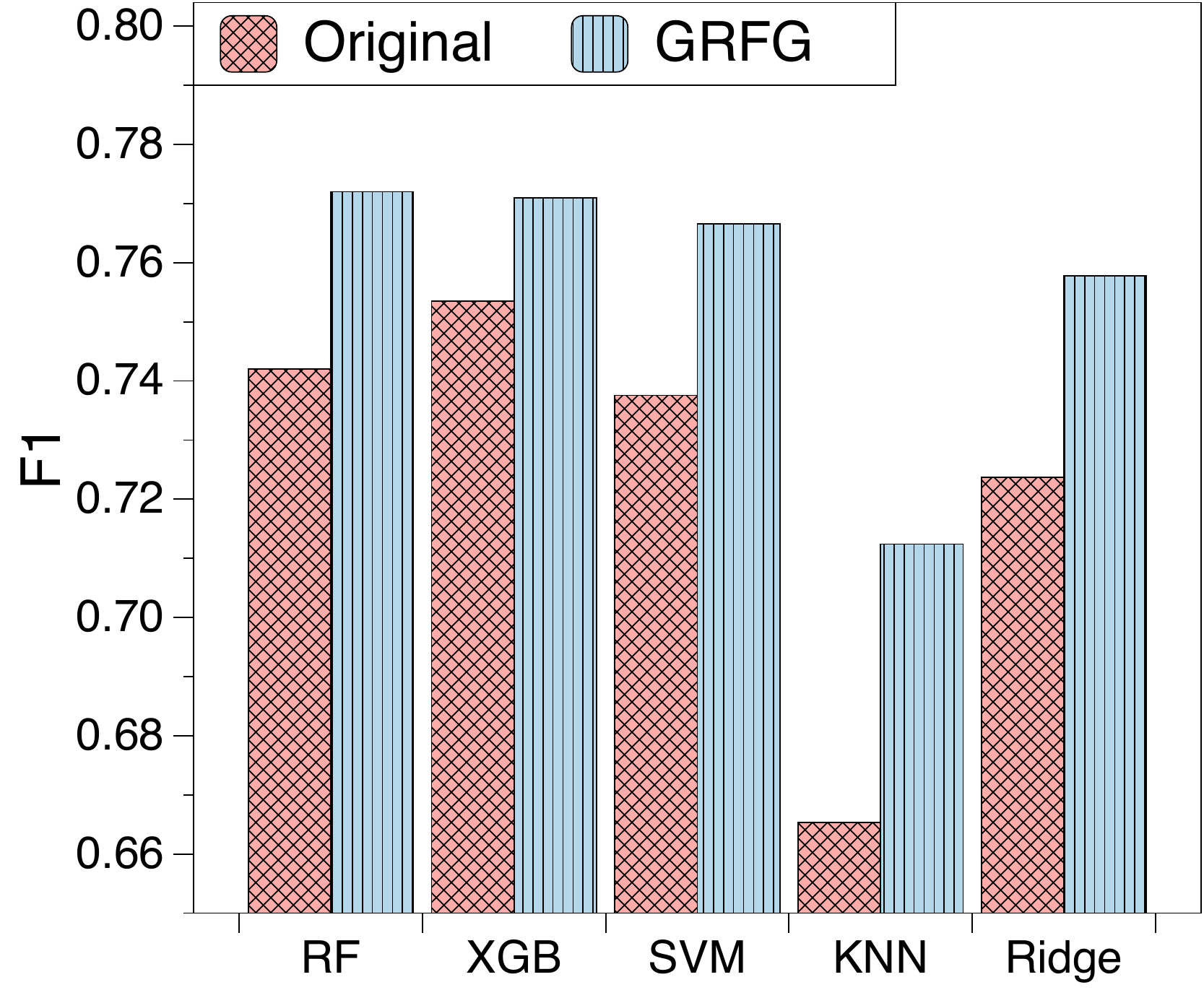}
}
\hspace{-3mm}
% % \vspace{-4mm}
\subfigure[Housing Boston]{
\includegraphics[width=3.4cm]{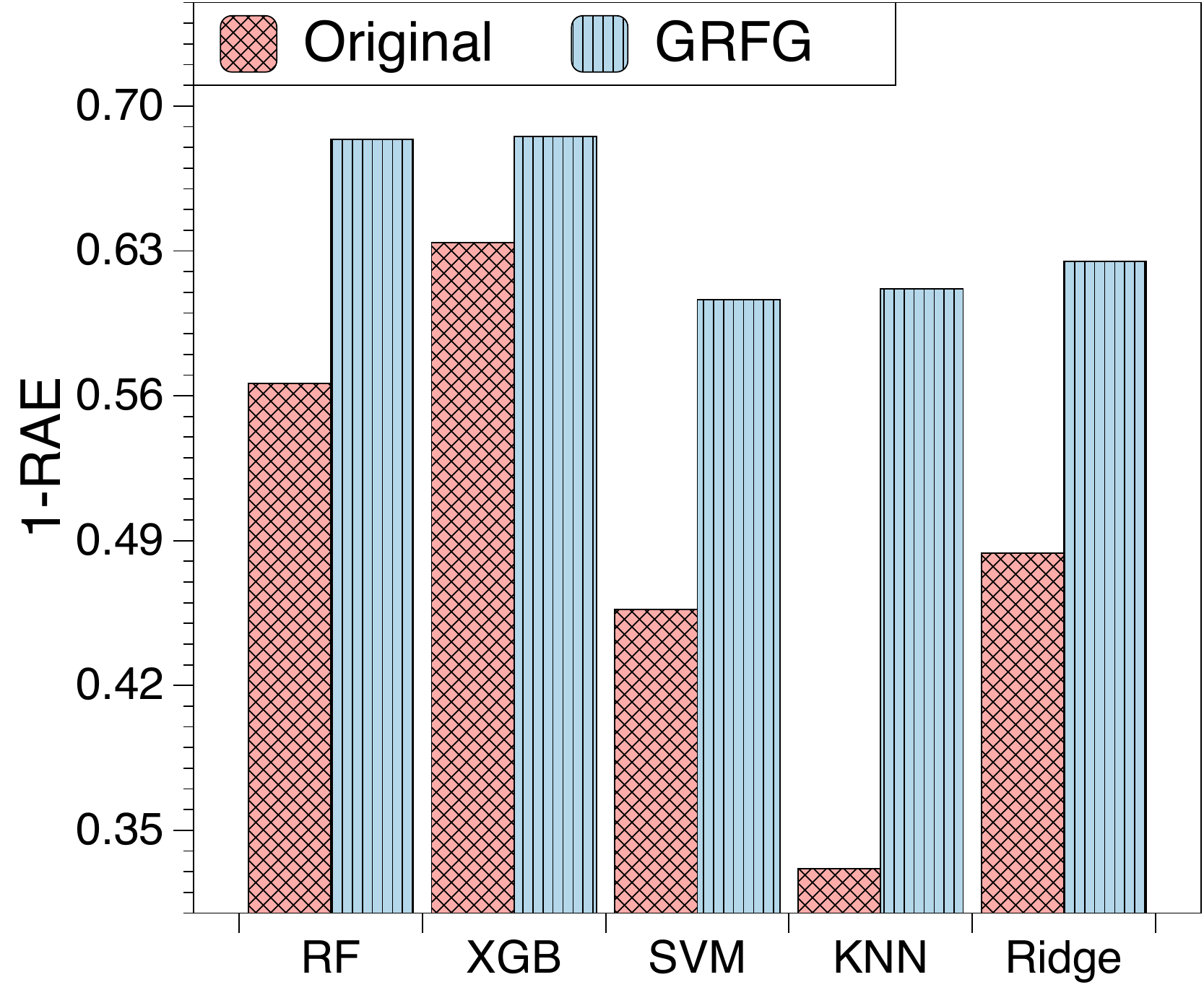}
}
\hspace{-3mm}
\subfigure[Openml 589]{ 
\includegraphics[width=3.4cm]{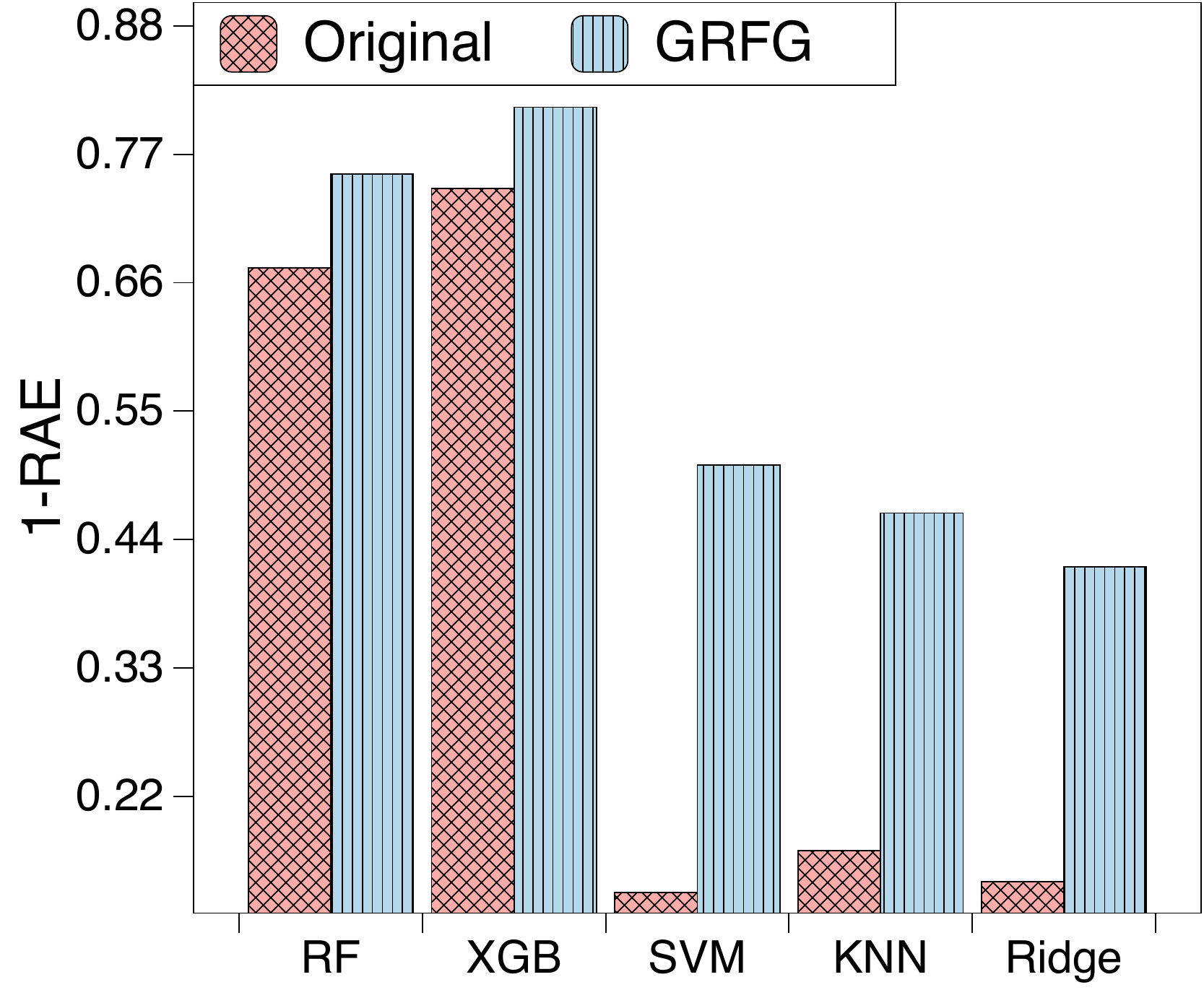}
}
\vspace{-0.35cm}
\caption{Comparison of different machine learning models in terms of F1 or 1-RAE.}
\label{differ_ml}
\vspace{-0.4cm}
\end{figure*}

This experiment is to answer:
\textit{Is GRFG robust when different ML models are used as a downstream task?}
We examined the robustness of GRFG by changing the ML model of a downstream task to Random Forest (RF), Xgboost (XGB), SVM, KNN, and Ridge Regression, respectively.
Figure ~\ref{differ_ml} shows the comparison results in terms of F1 score or 1-RAE on the datasets used in the Section \ref{study_lp}.
We observed that GRFG robustly improves model performances regardless of the ML model used.
This observation indicates that GRFG can generalize well to various benchmark applications and ML models.
We found that  RF and XGB are the two most powerful and robust predictors, which is consistent with the finding in Kaggle.COM competition community. 
Intuitively,  the accuracy of RF and XGB usually represent the performance ceiling on modeling a dataset. 
% It is hard to break the performance ceiling.
But, after using our method to reconstruct the data, we continue to significantly improve the accuracy of RF and XGB and break through the performance ceiling.   
This finding clearly validates the strong robustness of our method. 

\begin{figure}[!t]
\centering
\subfigure[Original Feature Space]{
\includegraphics[width=4.0cm]{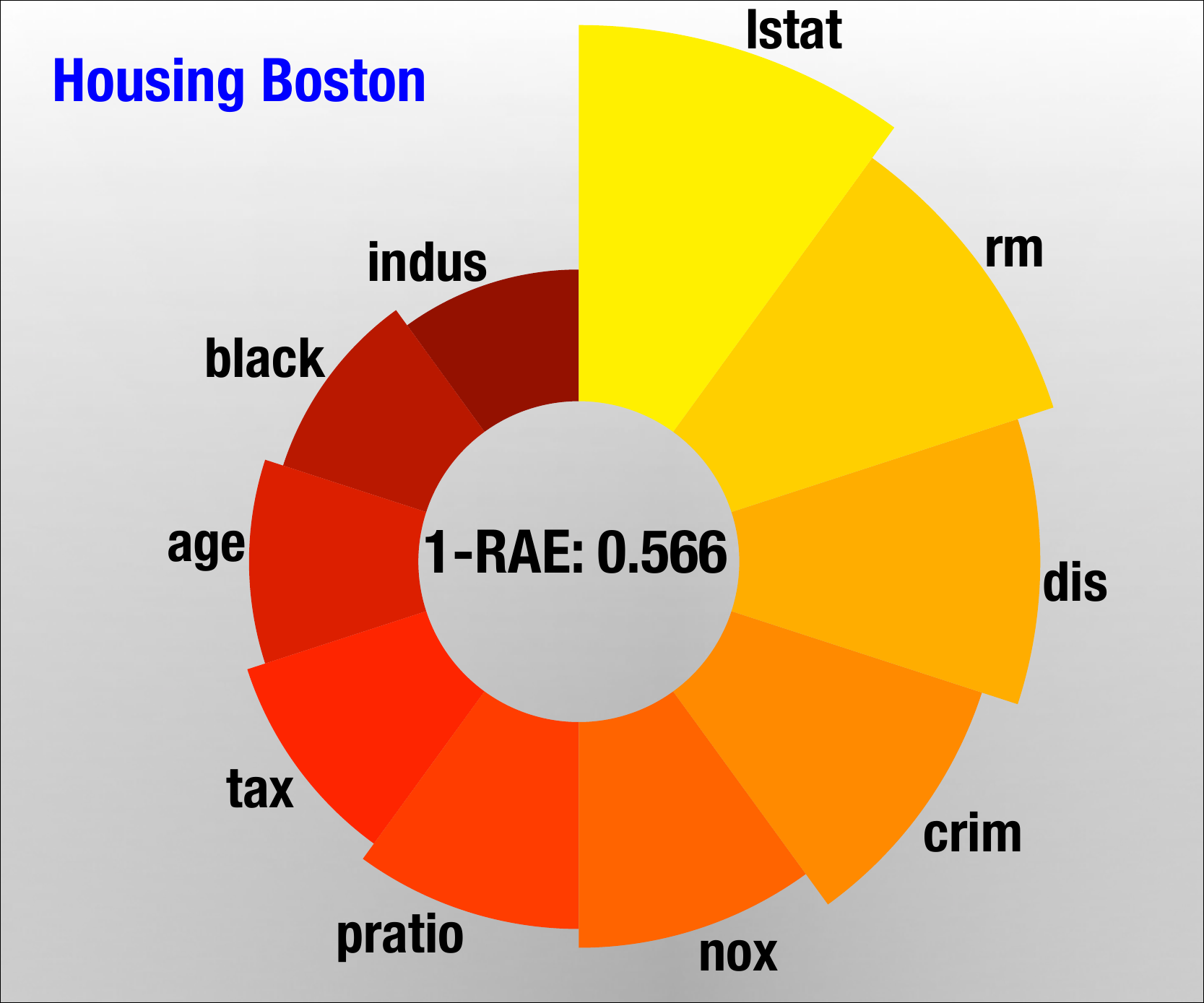}
}
\hspace{0mm}
\subfigure[Reconstructed Feature Space]{ 
\includegraphics[width=4.0cm]{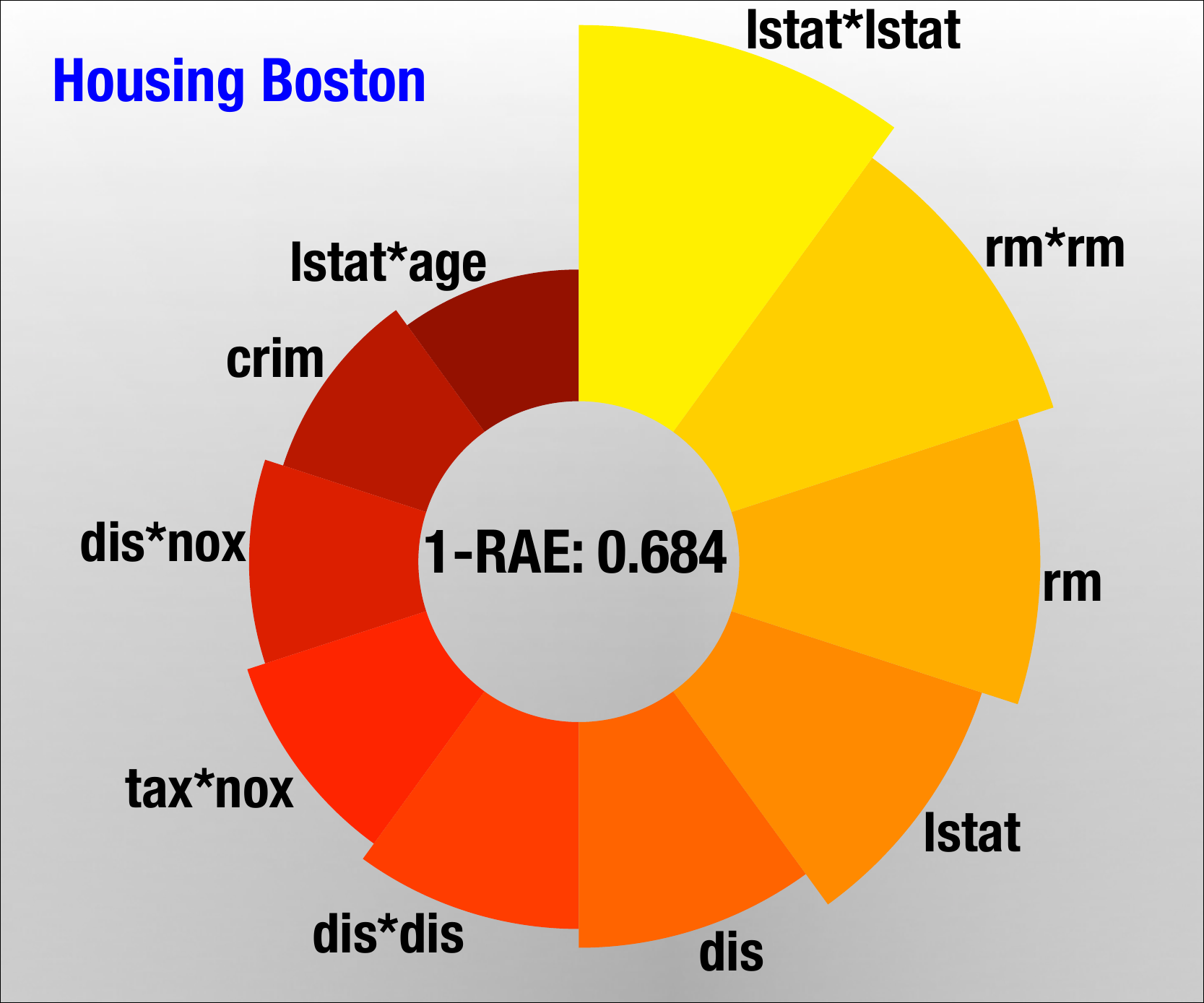}
}
\vspace{-0.35cm}
\caption{Top10 features for prediction in the original and GRFG-reconstructed feature space.}
\label{explain}
\vspace{-0.5cm}
\end{figure}

\subsubsection{Study of the traceability and explainability of GRFG}
This experiment aims to answer: \textit{Can GRFG generate an explainable feature space? Is this generation process traceable?}
We identified the top 10 essential features for prediction in both the original and reconstructed feature space using the Housing Boston dataset to predict housing prices with random forest regression.
Figure ~\ref{explain} shows the model performances in the central parts of each sub-figure. 
The texts associated with each pie chart describe the feature name.
If the feature name does not include an operation, the corresponding feature is original; otherwise, it is a generated feature. 
The larger the pie area is, the more essential the corresponding feature is.
We observed that the GRFG-reconstructed feature space greatly enhances the model performance by 20.9$\%$ and the generated features cover 60$\%$ of the top 10 features.
This indicates that GRFG generates informative features to refine the feature space.
Moreover, we can explicitly trace and explain the source and effect of a feature by checking its name.
For instance, ``lstat'' measures the percentage of the lower status populations in a house, which is negatively related to housing prices.
The most essential feature in the reconstructed feature space is 
``lstat*lstat'' that is generated by applying a ``multiply'' operation to ``lstat''.
This shows the generation process is traceable and the relationship between ``lstat'' and  housing prices is non-linear.
%Domain experts can undertake more in-depth analysis using these explicit and traceable feature names.

% {\color{blue} \subsubsection{Discussion of the training time and limitations of GRFG}
% This experiment aims to answer: \textit{What is the major limitations of GRFG?}
% }

%% file: related.tex
% \vspace{-0.1cm}
\section{Related Works}
\noindent\textbf{Reinforcement Learning (RL)} is the study of how intelligent agents should act in a given environment in order to maximize the expectation of cumulative rewards~\cite{sutton2018reinforcement}.
According to the learned policy, we may classify reinforcement learning algorithms into two categories: value-based and policy-based.
Value-based algorithms (\textit{e.g.} DQN~\cite{mnih2013playing}, Double DQN~\cite{van2016deep}) estimate the value of the state or state-action pair for action selection.
Policy-based algorithms (\textit{e.g.} PG~\cite{sutton2000policy}) learn a probability distribution to map state to action for action selection.
Additionally, an actor-critic reinforcement learning framework is proposed to  incorporate the advantages of value-based and policy-based algorithms~\cite{schulman2017proximal}.
In recent years, RL has been applied to many domains (e.g. spatial-temporal data mining, recommended systems) and achieves great achievements~\cite{wang2022reinforced,wang2022multi}.
In this paper, we formulate the selection of feature groups and operation as MDPs and propose a new cascading agent structure to resolve these MDPs.
% Our framework's simple model structure enables it to be applied in many scenarios easily and flexibly.

\noindent\textbf{Automated Feature Engineering} aims to enhance the feature space through feature generation and feature selection in order to improve the performance of machine learning models~\cite{chen2021techniques}.
Feature selection is to remove redundant features and retain important ones, whereas feature generation is to create and add meaningful variables. 
% Automated feature engineering  can be achieved by combining feature generation and feature selection with a certain algorithmic structure.
% yang1997comparative,hall1999feature, yang1998feature,narendra1977branch,
\ul{\textit{Feature Selection}} approaches include:
(i) filter methods (\textit{e.g}., univariate selection \cite{forman2003extensive}, correlation based selection \cite{yu2003feature}), in which features are ranked by a specific score like redundancy, relevance;  (ii) wrapper methods (\textit{e.g.}, Reinforcement Learning~\cite{ liu2021efficient}, Branch and Bound~\cite{ kohavi1997wrappers}), in which the optimized feature subset is identified by a search strategy under a predictive task;  (iii) embedded methods (\textit{e.g.}, LASSO \cite{tibshirani1996regression}, decision tree \cite{sugumaran2007feature}), in which selection is part of the optimization objective of a predictive task. 
% latent dirichlet allocation~\cite{blei2003latent},
\ul{\textit{Feature Generation}} methods include: (i) latent representation learning based methods, e.g. deep factorization machine~\cite{guo2017deepfm}, deep representation learning~\cite{bengio2013representation}. 
Due to the latent feature space generated by these methods, it is hard to trace and explain the extraction process.
(ii) feature transformation based methods, which use arithmetic or aggregate operations to generate new features~\cite{khurana2018feature,chen2019neural,xiao2023traceable}.
These approaches have two weaknesses: (a) ignore feature-feature heterogeneity  among different feature pairs; (b) grow exponentially when the number of exploration steps increases.
Compared with prior literature, our personalized feature crossing strategy captures the feature distinctness,  cascading agents learn effective feature interaction policies, and  group-wise generation manner accelerates feature generation.

%% file: conclusion.tex
% \vspace{-0.2cm}
\section{Conclusion Remarks}
% {\color{green} Rewriting the conclusion part according to the abstract and introduction parts. Highlights the journal version work and contributions and simplifies the description in the conference version. Meanwhile, modify the cover letter according to the introduction part.}
We present a group-wise reinforcement feature transformation framework for optimal and traceable representation space reconstruction to  improve the performances of predictive models. 
This framework nests feature generation and selection in order to iteratively reconstruct a recognizable and size-controllable feature space via feature-crossing.
To efficiently refine the feature space, we suggest a group-wise feature transformation mechanism and propose a new feature clustering algorithm (M-Clustering).
In this journal version, besides exploring more forms of downstream tasks, we enhance the framework from two perspectives: 
% (1) presenting an enhanced state representation approach, which improves the comprehension of reinforced agents for the current feature set, resulting in more effective policy learning.
% {\color{red} (2) proposing a new training strategy for reinforced agents, which addresses Q-value overestimation in them by decoupling the max operation in Q-learning, leading to rapid convergence and robust policy learning. }
(1) reformulating the original framework from the perspectives of the advanced state representation method and the improved Q-learning strategy.
(2) proposing an Actor-Critic optimization perspective to re-design the original feature transformation framework. 
We conducted an enlightening evaluation in a comprehensive set of experiments, scrutinizing our proposed methodologies from two distinct perspectives. 
We found that introducing an advanced state representation method gave the reinforcement learning agents a more precise understanding of the manipulated feature set, thus enhancing the model performance.
Besides, the improving Q-learning training strategy notably bolstered the robustness of the model's convergence process. This enhancement facilitated a faster convergence rate, despite the observed performance remaining commensurate with the original model's.
Most crucially, our experiments revealed that utilizing the Actor-Critic framework yielded significant benefits. 
This approach consistently demonstrated superior convergence efficiency, thereby expediting the learning process.
Moreover, the Actor-Critic paradigm demonstrated substantial superiority in the context of downstream task performance, thereby highlighting its significant potential within reinforcement learning-based feature transformation tasks.
% Through extensive experiments, we can find that human intelligence in feature transformation can be modeled by reinforced agents.
% Group-wise feature transformation mechanism can efficiently explore feature space and augment reward signals of reinforced agents to learn an optimal feature space.
% Improving the learning and comprehension of reinforced agents can expedite and intensify the search for the optimal feature space.
In the future, we aim to include the pre-training technique in GRFG  to further enhance feature transformation.

%% file: acknowledgement.tex
\begin{acks}

This work is partially supported by IIS-2152030, IIS-2045567, and IIS-2006889.

\end{acks}